%% file: main.tex
\documentclass[11pt]{article}

\usepackage[final]{acl}

\usepackage{times}
\usepackage{latexsym}

\usepackage[T1]{fontenc}
\usepackage[utf8]{inputenc}
\usepackage{microtype}
\usepackage{inconsolata}

\usepackage{graphicx}
\usepackage{booktabs}   % for \toprule, \midrule, \bottomrule
\usepackage{amssymb}    % for \checkmark
\usepackage{tabularx}   % for auto-width tables
\usepackage{multirow}   % for vertical cell spanning
\usepackage{pifont}     % for \cmark and \xmark
\usepackage{colortbl}   % for \rowcolor in tables
\newcommand{\cmark}{\ding{51}}
\newcommand{\xmark}{\ding{55}}
\newcommand{\our}{RULER}

\usepackage{amsmath}
\usepackage{amssymb}
\usepackage{enumitem}

\graphicspath{{figures/}{tables/}}

\title{\our: Instance-aware Rubric Rewards for SVG Generation}

\usepackage[most]{tcolorbox}
\usepackage{xcolor}

\newtcolorbox{prompt}[1]{
  breakable,
  enhanced,
  colback=gray!4,
  colframe=gray!45,
  title=\textbf{#1},
  fonttitle=\small,
  fontupper=\scriptsize,
  boxrule=0.4pt,
  arc=1.5pt,
  left=4pt,
  right=4pt,
  top=4pt,
  bottom=4pt,
  before skip=6pt,
  after skip=6pt
}

\author{
Hangyu Ran\textsuperscript{1,2,*}
\quad
Yuhao Zheng\textsuperscript{3,*}
\quad
Yingying Zhang\textsuperscript{1}
\quad
Kevin Qinghong Lin\textsuperscript{4,$\dagger$}
\quad
Han Peng\textsuperscript{1,$\dagger$}
\\
\textsuperscript{1}Ant Group
\qquad
\textsuperscript{2}The Hong Kong University of Science and Technology (Guangzhou)
\\
\textsuperscript{3}Independent Researcher
\qquad
\textsuperscript{4}University of Oxford
\\[-0.1em]
{\small
\textsuperscript{*}Equal contribution
\qquad
\textsuperscript{$\dagger$}Corresponding authors
}
}

\begin{document}
\maketitle
% \raggedbottom
% ============================================================
% Sections
% ============================================================
\input{sections/abstract}
\input{sections/intro}
\input{sections/related_work}
\input{sections/method}
\input{sections/experiment}
\input{sections/conclusion}

% ============================================================
% Limitations (required for *ACL submissions)
% ============================================================

\input{sections/limitations}

% Acknowledgments
% ============================================================
% \section*{Acknowledgments}

\input{sections/acknowledgement}
% Bibliography
\bibliography{references}

\appendix

\input{sections/appendix}

\end{document}

%% file: sections/abstract.tex
\begin{abstract}
Generating Scalable Vector Graphics (SVG) code from natural-language instructions is an open-ended task without absolute visual ground truth, leaving both evaluation and policy optimization without a faithful signal. Scalar metrics (CLIP, Aesthetic) calibrated on natural images transfer poorly to stylized vector content, and reusing them as RL rewards triggers reward hacking. We address both limitations with rubric-based scoring. We first establish empirically that prompting a vision--language judge with a multi-axis rubric correlates with human judgments far better than scalar metrics, both across samples and within instructions. Building on this finding, we introduce \textbf{\our{}} (Instance-aware \textbf{Ru}bric Rewards for Reinforcement \textbf{LE}a\textbf{R}ning), which converts each instruction into an \emph{instance-aware rubric} of six items spanning semantic, visual, and stylistic axes; a judge VLM scores rendered rollouts item-by-item, and the weighted satisfactions form a fine-grained reward optimized via Group Relative Policy Optimization. Because the rubric is derived from text alone, \our{} requires neither paired SVG ground truth nor human preference labels. On MMSVG-Illustration and MMSVG-Icon, \our{} lifts the rubric score from $0.432/0.395$ to $0.693/0.683$, surpassing dedicated SVG specialists and matching the substantially larger DeepSeek-V3, with ablations identifying rubric design as the active lever for RL on open-ended SVG generation. The project page is available at \url{https://hangyuran.github.io/RULER/}.
\end{abstract}

%% file: sections/intro.tex
\section{Introduction}
\label{sec:intro}

Generating Scalable Vector Graphics (SVG) ~\cite{starvector,omnisvg,svgen,deepsvg} has emerged as a crucial frontier in visual code generation. As a unique form of text that renders into precise graphics, SVG code is structured, executable and controllable, unlike descriptive natural language ~\cite{vcode,code2world,code2video}. Owing to these distinctive properties, frontier foundation models increasingly prioritize SVG generation to showcase their visual code synthesis capabilities ~\cite{kimi-k25,gemini2.5, gpt5}. This widespread interest has led to pioneering efforts in dataset curation ~\cite{omnisvg, unisvg}, benchmarking ~\cite{vcode}, and specialized model training paradigms ~\cite{sgp,rensonsvg,render-aware,vfig}. 

\begin{figure*}[t]
\vspace{-10pt}
  \centering
  \includegraphics[width=\textwidth]{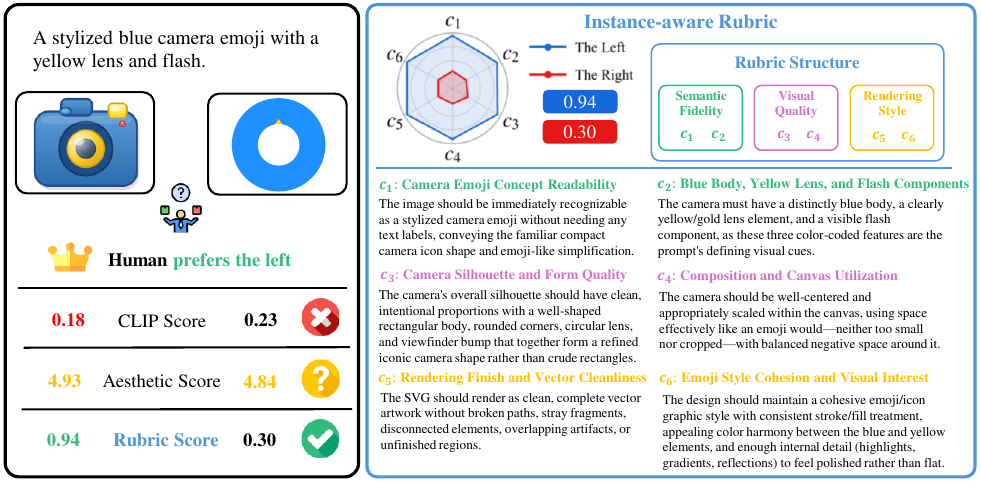}
\caption{\textbf{Comparison of evaluation metrics and the RULER rubric.} \textbf{Left:} Standard scalar metrics (CLIP and Aesthetic) mis-rank or fail to penalize a broken SVG, whereas our Rubric score aligns with human preference. \textbf{Right:} The structured six-item instance-aware rubric spanning semantic, visual, and stylistic axes.}
  \label{fig:teaser}
  \vspace{-18pt}
\end{figure*}

% foundamental challenge
Within this landscape, generating structured SVG code directly from natural instructions stands as a fundamental research challenge. This core difficulty stems from an inherent characteristic of open-ended synthesis: a single instruction can map to countless semantically valid renderings, leaving no absolute visual ground truth to serve as a standard reference. Consequently, the field is bottlenecked on two closely coupled fronts:
\begin{itemize}[leftmargin=*, itemsep=3pt, topsep=2pt]
\item \textbf{Unreliable metrics for evaluation.} CLIPScore ~\cite{clipscore}, aesthetic classifiers, and the Human Preference Score ~\cite{hps} were calibrated on photorealistic natural images and transfer poorly to stylized vector content. As shown in Figure~\ref{fig:teaser}, they frequently assign higher scores to broken SVGs than to faithful ones, and correlate weakly with human judgment both within and across models.
\item \textbf{Unfaithful supervision signal for training.} Supervised fine-tuning on instruction--SVG pairs reduces to behavioral cloning of dataset-specific templates and fails to generalize ~\cite{starvector,omnisvg}. Reinforcement learning is the natural alternative, but its effectiveness is dominated by the choice of reward ~\cite{effects,uimate}. The rewards available for SVG generation are exactly the unreliable metrics above, so the policy drifts toward whichever signal is easiest to inflate rather than toward better generations~\cite{skalse2022defining,gao2023scaling}.
\end{itemize}

% our method
To address these limitations, \textbf{\textit{(i)}} \textbf{for evaluation}, 
we first conduct a systematic empirical analysis (detailed in Section \ref{sec:method}) utilizing 900 human-annotated SVG samples generated by models of varying capabilities. We observe that \textit{rubric-based scoring}, which prompts a vision-language judge to rate rendered SVGs along multiple decomposed axes, correlates with human preference far better than conventional scalar metrics. Specifically, it achieves a sample-level correlation (Spearman's $\rho$) of 0.7929 and a pairwise ranking agreement (Goodman-Kruskal $\gamma$) of 0.7574, establishing it as a robust evaluator for open-ended SVG quality.
\textbf{\textit{(ii)}} \textbf{For training}, to provide the policy with fine-grained supervision, we introduce \textbf{\our{}} (Instance-aware \textbf{Ru}bric Rewards for Reinforcement \textbf{LE}a\textbf{R}ning). By repurposing our robust evaluation mechanism into a reward signal, \our{} converts each instruction into an \emph{instance-aware rubric} of six items spanning \emph{semantic fidelity}, \emph{visual quality}, and \emph{rendering style}. A judge VLM then scores each rendered rollout item-by-item, and the weighted satisfactions form a fine-grained reward optimized via Group Relative Policy Optimization. Because the rubric is generated from text alone, \our{} requires neither paired SVG ground truth nor human preference labels and \textit{scales to any unannotated instruction set}.

On MMSVG-Illustration and MMSVG-Icon ~\cite{omnisvg}, \our{} lifts the rubric score from $0.432/0.395$ (the Qwen3-8B ~\cite{qwen3} backbone) to $0.693/0.683$, surpassing dedicated SVG specialists and matching the substantially larger DeepSeek-V3 ~\cite{deepseek-v3}. Ablations further identify rubric design as the active lever for RL on open-ended visual code. Our contributions are threefold:

\begin{itemize}[leftmargin=*, topsep=4pt, partopsep=0pt, itemsep=2pt, parsep=0pt]

\item \textbf{Rubric-based Evaluation for SVG Quality.}
We systematically investigate the evaluation paradigm of SVG generation, exposing the severe insensitivity of standard scalar metrics to actual visual quality under domain shift. To address this, we adopt a \emph{rubric-based} score and show that it correlates substantially better with human preference. 

\item \textbf{Instance-aware Rubric for Learning.}
Building on this analysis, we propose \textbf{\our{}}, which generates a six-item rubric per instruction spanning semantic, visual, and stylistic axes, and uses it as a dense, query-conditioned reward for RL, turning ambiguous visual judgments into explicitly verifiable sub-goals.

\item \textbf{State-of-the-Art Performance.}
Extensive experiments on MMSVG-Illustration and MMSVG-Icon show that \our{} achieves the strongest Rubric scores, surpassing dedicated SVG specialists and the substantially larger DeepSeek-V3 while remaining competitive on conventional metrics, and outperforms standard RL baselines.

\end{itemize}

%% file: sections/related_work.tex
\section{Related Work}
\label{sec:related_work}

\input{tables/reward_comparison}

\subsection{SVG Code Generation}
SVG code generation has progressed from optimization-based path tracing~\cite{diffvg,vectorfusion,svgdreamer} to autoregressive primitive-aware generation~\cite{omnisvg, unisvg, starvector}, and most recently to rendering-aware reinforcement learning~\cite{render-aware} that optimizes the policy directly against the rendered image. A central but unresolved question across these pipelines is how to score a generated SVG when no paired ground truth exists; prior reward designs (Table~\ref{tab:reward_comparison}) each fall short on at least one desideratum. \textit{Pixel-based} metrics (SSIM, PSNR) require a paired reference and reduce quality to a single scalar; \textit{rule-based} signals such as code length~\cite{render-aware} score only the code without inspecting the rendered image; \textit{embedding} scores like CLIPScore~\cite{clipscore} are reference-free and visually grounded, but provide only coarse-grained assessments of compositional correctness~\cite{geneval} and can be easy to game under RL~\cite{render-aware}; and \textit{universal-rubric} scoring~\cite{vfig,vectorgym} provides multi-dimensional feedback yet ignores instruction-specific notions of correctness. Our \our{} closes this gap with an \textit{instance-aware rubric} that is simultaneously reference-free, multi-dimensional, visually grounded, and instance-conditioned.

\subsection{Rubric-Based Evaluation and Rewards}
Rubric-based evaluation decomposes quality into interpretable criteria without relying on auxiliary reward models. ~\citet{llm-rubric} introduce LLM-Rubric for calibrated multi-aspect evaluation. Building on this evaluation primitive, RaR~\cite{rar} and RGR-GRPO~\cite{rgr-grpo} use rubric scores directly as RL rewards, showing that fine-grained checklist feedback can improve policy training on text-domain tasks such as instruction following and reasoning. 
Our \our{} extends rubric-based RL in two complementary directions: it applies rubric-based rewards to open-ended \emph{visual code} through a VLM judge that scores rendered SVGs against instance-aware rubrics, and it constructs these rubrics without ground-truth SVGs, providing case-specific supervision without restricting generation to a single reference SVG.

%% file: tables/reward_comparison.tex
\begin{table*}[t]
\centering
\caption{\textbf{Comparison of reward paradigms for SVG generation.} 
We evaluate existing approaches across five critical desiderata. GT-free denotes that ground-truth visual references are not required.}
\label{tab:reward_comparison}
\renewcommand{\arraystretch}{1.2}
\newcommand{\gmark}{\textcolor{green!70!black}{\cmark}}
\newcommand{\rmark}{\textcolor{red!70!black}{\xmark}}
\resizebox{\textwidth}{!}{%
\begin{tabular}{@{} l l c c c c c @{}}
\toprule
\multirow{2}{*}{\textbf{Reward Paradigm}} 
& \multirow{2}{*}{\textbf{Representative Metric}} 
& \textbf{Visual} 
& \textbf{Semantic} 
& \multirow{2}{*}{\textbf{GT-Free}} 
& \multirow{2}{*}{\textbf{Multi-Axis}} 
& \textbf{Instance-} \\
& 
& \textbf{Granularity} 
& \textbf{Granularity} 
& 
& 
& \textbf{Aware} \\
\midrule
\textbf{Pixel-based} 
& SSIM / PSNR ~\cite{ssim}
& Fine-grained 
& N/A 
& \rmark 
& \rmark 
& \rmark \\

\textbf{Rule-based} 
& Code Length ~\cite{render-aware}
& N/A 
& N/A 
& \gmark 
& \rmark 
& \rmark \\

\textbf{Embedding-based} 
& CLIPScore ~\cite{clipscore}
& Coarse-grained 
& Coarse-grained 
& \gmark 
& \rmark 
& \rmark \\

\midrule

\textbf{Rubric-based} 
& Universal Rubric ~\cite{vfig}
& Fine-grained 
& Fine-grained 
& \rmark 
& \gmark 
& \rmark \\

\rowcolor{gray!10}
\textbf{Rubric-based} 
& \textbf{Instance-aware Rubric (\our{})} 
& \textbf{Fine-grained} 
& \textbf{Fine-grained} 
& \gmark 
& \gmark 
& \gmark \\
\bottomrule
\end{tabular}%
}
\end{table*}

%% file: sections/method.tex
\begin{figure*}[t]
\vspace{-8pt}
  \centering
  \includegraphics[width=0.9\textwidth]{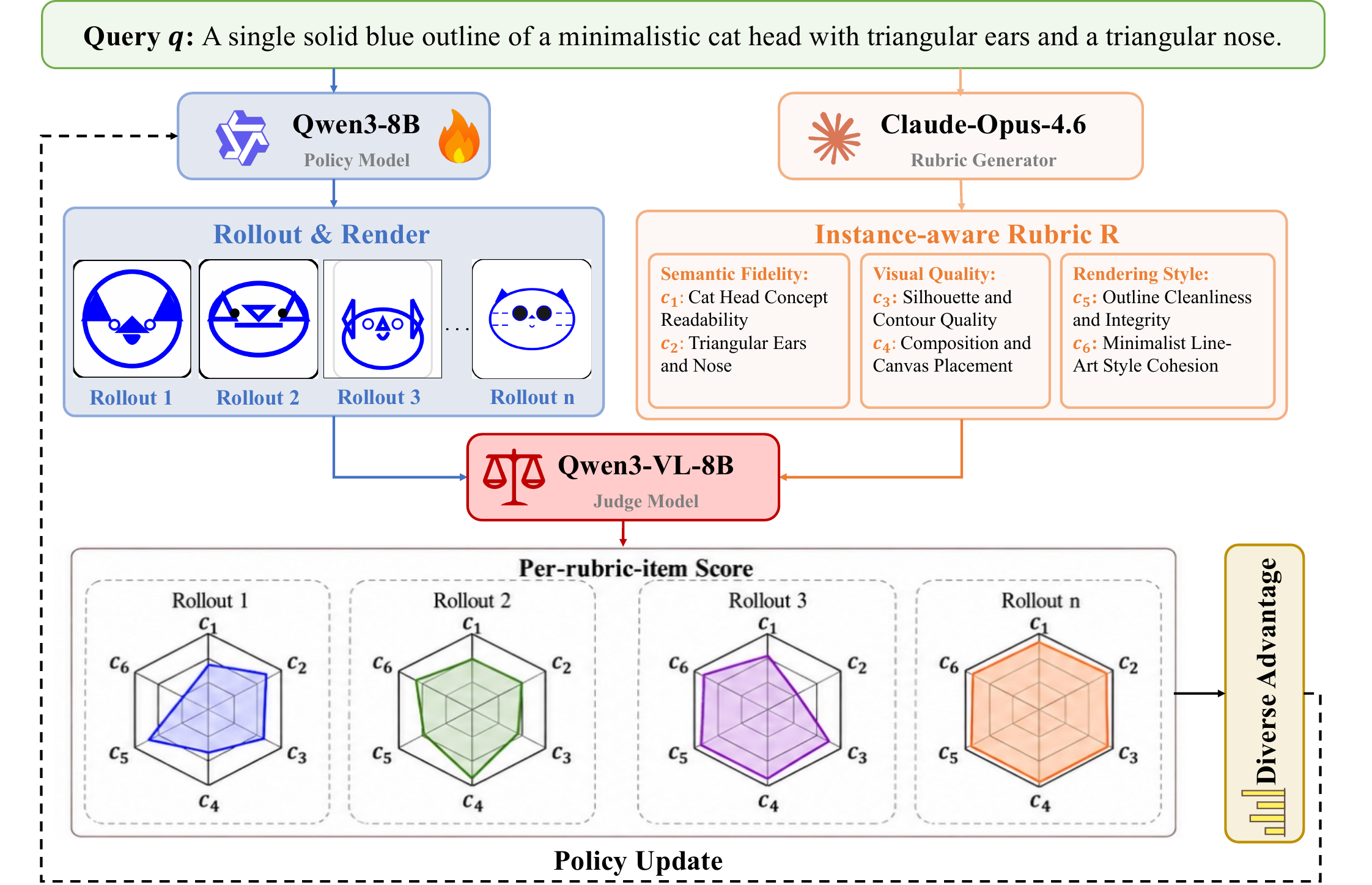}
  \caption{\textbf{Overview of \our{}.} A frontier model derives an instance-aware rubric---six items across semantic, visual, and stylistic axes---from each text instruction (left). The policy samples SVG rollouts that are rendered and scored by a judge VLM against the rubric; the weighted per-item satisfactions form the GRPO reward signal that drives policy updates (right).}
  \label{fig:framework}
  \vspace{-10pt}
\end{figure*}

\section{\our{}}
\label{sec:method}
In this section, we present \our{} as illustrated in Figure \ref{fig:framework}. We first formulate open-ended SVG generation as a token-level Markov Decision Process (\S\ref{subsec:task_def}), which highlights the design of the reward function as the central challenge of this task. To address this bottleneck, we conduct a systematic empirical analysis in \S\ref{subsec:rubric_eval} to establish rubric-based scoring as a robust and reliable evaluation primitive. Building on these empirical findings, we detail the core components of our framework: a scalable pipeline that constructs instance-aware rubrics from text instructions (\S\ref{subsec:rubric_gen}), and a reinforcement learning pipeline that optimizes the policy using these rubrics via Group Relative Policy Optimization (GRPO) (\S\ref{subsec:grpo}).

\subsection{Task Definition}
\label{subsec:task_def}
We formulate open-ended SVG generation as a token-level Markov Decision Process (MDP). Given an instruction $\mathcal{Q}$ that specifies the desired visual content, a language model policy $\pi_\theta$ autoregressively generates a structured SVG sequence $\mathcal{Y} = (y_1, \dots, y_T)$. At step $t$, the state $s_t = (\mathcal{Q}, y_{<t})$ concatenates the instruction with the prefix already generated, the action $a_t = y_t$ is the next token sampled from $\pi_\theta(\cdot \mid s_t)$, and the transition $s_{t+1} = s_t \oplus a_t$ is deterministic. Once the sequence terminates, a deterministic rendering engine $\mathcal{E}$ executes the completed code into a visual representation $\mathcal{I}_{\text{gen}} = \mathcal{E}(\mathcal{Y})$, on which the reward function $\mathcal{R}(\cdot)$ is computed. Because open-ended SVG generation lacks an absolute visual ground truth, the design of $\mathcal{R}(\cdot)$---rather than the optimization machinery---is the central question raised by this MDP. The objective is to learn the optimal $\theta$ that maximizes the expected reward.

\begin{figure}[t]
% \vspace{-10pt}
  \centering
  \includegraphics[width=\columnwidth]{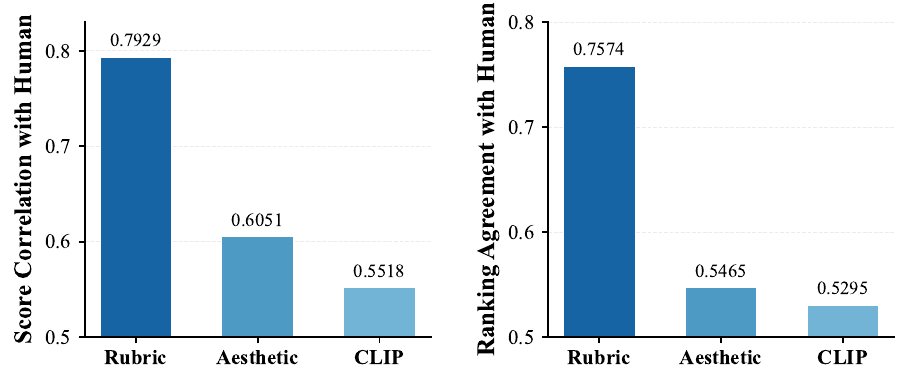}

  \caption{Rubric score shows superior human alignment compared with Aesthetic and CLIP.}

  \label{fig:human-study}
  \vspace{-18pt}
\end{figure}

% \subsection{Reliable Rubric-based Evaluator}
\subsection{Human Assessment of Existing Metrics}
\label{subsec:rubric_eval}
Before constructing the reward signal, we verify whether rubric-based scoring aligns with human judgment for stylized vector content. We collect 900 rendered SVG samples generated by three models of varying capabilities (\texttt{Claude-Opus-4.6}, \texttt{Qwen3-32B}, and \texttt{Qwen3-8B} ~\cite{qwen3}, 300 each) and \textit{obtain human quality ratings} following the annotation protocol detailed in Appendix~\ref{app:human_alignment_protocol}. We evaluate metrics against these annotations from two complementary perspectives: sample-level score correlation and pairwise ranking agreement.

\noindent\textbf{Score correlation.} 
As shown in Figure \ref{fig:human-study} (left), we compute the Spearman rank correlation (Spearman's $\rho$) between automated metrics and human scores over all 900 cases. Rubric-based scoring achieves a correlation of $\rho = 0.7929$, outperforming Aesthetic ($\rho = 0.6051$) and CLIP ($\rho = 0.5518$). This confirms that decomposing evaluation into explicit axes tracks human-perceived quality more effectively than conventional scalar metrics.

\noindent\textbf{Pairwise ranking agreement.} 
To assess ranking stability, we measure agreement using the \textit{Goodman-Kruskal Gamma ($\gamma$)} statistic. For each of the 300 evaluation triples (comprising 900 total pairs across the three generators), we calculate the consistency of metric-induced pairwise orderings against human preferences. 
As shown in Figure \ref{fig:human-study} (right), the rubric-based evaluator achieves a strong directional agreement of $\gamma = 0.7574$. This substantially exceeds Aesthetic ($\gamma = 0.5465$) and CLIP ($\gamma = 0.5295$). Gamma assesses consistency across all possible pairwise comparisons within each triple, indicating that the rubric serves as a highly reliable evaluator that \textit{aligns closely with human preferences}.

\subsection{Instance-Aware Rubric Generation}
\label{subsec:rubric_gen}

Designing a faithful reward $\mathcal{R}(\cdot)$ without paired ground truth is the central question raised by the MDP above. As established in Section \ref{subsec:rubric_eval}, scalar metrics (CLIP, Aesthetic, HPS) compress multi-dimensional visual quality into an opaque score and are unreliable on stylized vector content. To bypass this, \our{} elicits tailored evaluation criteria from a frontier model $\mathcal{M}_{\text{rub}}$ (\textit{e.g.,} \texttt{Claude-Opus-4.6} ~\cite{opus46}) using only the unannotated instruction $\mathcal{Q}$. We prompt $\mathcal{M}_{\text{rub}}$ to produce a discrete, instance-aware rubric of six items grouped along three complementary axes. To preserve the open-ended solution space and prevent the rubric from degenerating into a reconstruction checklist, items are specified at the level of design intentions rather than exact pixel or path constraints. Each item targets a single observable axis, is independently judgeable from the rendered image, and penalizes a distinct type of failure, so that the rubric covers the multi-dimensional notion of visual quality without redundancy:
\begin{itemize}[leftmargin=*, itemsep=2pt, topsep=2pt]
\item \textbf{\textit{Semantic Fidelity}}: high-level concept readability and the visual presence of major components, distinctive cues, and prompt-specific relations.
\item \textbf{\textit{Visual Quality}}: silhouette and form refinement together with composition and canvas design.
\item \textbf{\textit{Rendering Style}}: rendering finish and execution cleanliness coupled with style cohesion and designed visual interest.
\end{itemize}
The full instance-aware descriptions, weighting scheme, and prompting templates are deferred to Appendix~\ref{app:prompt_templates}. The rubric is formally defined as
\begin{equation}
C_\mathcal{Q} = \{(c_k, w_k)\}_{k=1}^{6},
\end{equation}
where $c_k$ encapsulates the instance-aware title, description, and continuous scoring guide for item $k$ adapted to $\mathcal{Q}$, and $w_k$ is its importance weight.

\subsection{Policy Optimization}
\label{subsec:grpo}

\paragraph{Reward Calculation.}
At each RL step, the rendered image $\mathcal{I}_{\text{gen}} = \mathcal{E}(\mathcal{Y})$ is evaluated by a judge VLM $\mathcal{M}_{\text{judge}}$. Instead of querying for a holistic scalar score, we prompt $\mathcal{M}_{\text{judge}}$ to follow the rubric $C_\mathcal{Q}$ and independently rate $\mathcal{I}_{\text{gen}}$ on each item $c_k$, producing a continuous satisfaction $s_k \in [0,1]$ guided by an explicit scoring guide. The reward is computed as the normalized weighted average:
\begin{equation}
\mathcal{R}(\mathcal{I}_{\text{gen}}) =
\frac{\sum_{k=1}^{6} w_k s_k}
{\sum_{k=1}^{6} w_k}.
\end{equation}
This yields a dense, multi-dimensional signal in place of an opaque scalar.

\paragraph{Group Relative Advantage.}
The instance-aware rubric provides fine-grained, multi-axis feedback for each instruction (Figure~\ref{fig:framework}), so different rollouts in the same group often succeed unevenly across items and yield naturally diverse reward signals. 
This within-group diversity is precisely what Group Relative Policy Optimization (GRPO)~\cite{deepseekmath,deepseek-r1} exploits: GRPO estimates advantages from the relative rewards of outputs within each group, making it a natural fit for our reward structure.
For each instruction $\mathcal{Q}$, we sample $G$ rollouts $\{\mathcal{Y}_i\}_{i=1}^{G}$ from $\pi_{\theta_{\text{old}}}$, render each into $\mathcal{I}_i = \mathcal{E}(\mathcal{Y}_i)$, score it as $\mathcal{R}_i = \mathcal{R}(\mathcal{I}_i)$, and form group-normalized advantages
\begin{equation}
A_i = \frac{\mathcal{R}_i - \mathrm{mean}(\{\mathcal{R}_j\}_{j=1}^{G})}{\mathrm{std}(\{\mathcal{R}_j\}_{j=1}^{G}) + \epsilon}.
\end{equation}
Parameters are then updated by maximizing the clipped surrogate objective~\cite{schulman2017proximal}:
\begin{equation}
\small
\begin{aligned}
\mathcal{L}_{\text{GRPO}}(\theta) = \mathbb{E}_{\mathcal{Q}} \biggl[ \frac{1}{G} \sum_{i=1}^{G} \min \bigl( \rho_i A_i,\; \text{clip}(\rho_i, 1-\epsilon, \\ 1+\epsilon) A_i \bigr)],
\end{aligned}
\label{eq:grpo}
\end{equation}
where $\rho_i = \pi_\theta(\mathcal{Y}_i \mid \mathcal{Q}) / \pi_{\theta_{\text{old}}}(\mathcal{Y}_i \mid \mathcal{Q})$ is the sequence-level importance ratio. Through this process, \our{} iteratively refines its policy to maximize satisfactions across semantic fidelity, visual quality, and rendering style.

%% file: sections/experiment.tex
\section{Experiments}
\label{sec:experiment}
We structure our experimental analysis to answer the following research questions:
\textbf{RQ1:} How does \our{} compare against baselines on open-ended visual code generation?
\textbf{RQ2:} Does an instance-aware rubric reward mechanism outperform existing RL paradigms?
\textbf{RQ3:} How do the individual evaluation items and dimensions within the instance-aware rubrics contribute to the overall generation quality?
\textbf{RQ4:} How robust is \our{} across different base models and rubric generators?
\textbf{RQ5:} What qualitative differences emerge between \our{} and existing baselines on representative generation cases?
\input{tables/main_table}

\subsection{Experimental Setup}
\label{sec:exp_setup}

\paragraph{Baselines.}
We compare \our{} with three families of baselines: \textbf{\textit{(i)}} Diffusion-optimized methods, including VectorFusion and SVGDreamer; \textbf{\textit{(ii)}} Foundation LLMs, including Qwen3-8B and Qwen3-32B and the much larger DeepSeek-V3; and \textbf{\textit{(iii)}} SVG specialist models, including IconShop, JanusCoder-8B and OmniSVG-8B. These baselines cover diverse modeling paradigms, model scales, and architectural designs.

\paragraph{Benchmarks.}
We evaluate on two benchmarks: MMSVG-Illustration for richer illustrative content and MMSVG-Icon for compact icon-style generation ~\cite{omnisvg}.

\paragraph{Metrics.}
Following prior work, we report tokens per sample (efficiency), CLIP Score (text--image alignment), Aesthetic Score (an aesthetic classifier score), and the Human Preference Score (HPS), keeping these conventional metrics for comparability despite their known \textit{limitations on evaluation} (Figure~\ref{fig:teaser}). To capture holistic visual quality, we additionally introduce a \emph{Rubric score} that prompts GPT-5-mini ~\cite{gpt5} as an independent VLM-as-Judge to rate each rendered SVG against a shared universal rubric; we treat this Rubric score as the primary indicator of overall quality.

\noindent More implementation details are in Appendix \ref{app:training_details}.

\subsection{Main Results (RQ1)}
\label{sec:main_results}

Table~\ref{tab:main_results} reports the comparison against the three baseline families on two benchmarks. We summarize the main observations below.

\paragraph{Consistent improvements across baselines.}
\our{} outperforms every baseline family on the primary Rubric metric across both benchmarks. 
Against optimization-based methods such as VectorFusion and SVGDreamer, it achieves higher universal Rubric scores while generating SVG code end-to-end without per-prompt iterative optimization.
Against dedicated SVG specialists such as OmniSVG, IconShop and JanusCoder~\cite{omnisvg,iconshop,januscoder}, it lifts the universal Rubric score from $0.390$ to $0.693$ on Illustration and from $0.586$ to $0.683$ on Icon, indicating that closing the open-ended quality gap requires more than scaling SVG-specific pretraining. Among open-source foundation LLMs of comparable scale, \our{} clearly outperforms its own Qwen3-8B backbone (Rubric $0.432\!\rightarrow\!0.693$ on Illustration, $0.395\!\rightarrow\!0.683$ on Icon) and reaches visual quality on par with the much larger DeepSeek-V3, demonstrating that an 8B model trained with instance-aware rubric rewards can rival models at a substantially larger scale.
\paragraph{Competitive performance on auxiliary metrics, top-tier on the rubric score.}
Although we argue in Section \ref{subsec:rubric_eval} that CLIP and aesthetic scores are individually insufficient for evaluating open-ended SVG code generation, \our{} nevertheless achieves competitive CLIP, Aesthetic, and HPS scores across both benchmarks, ruling out the concern that our rubric gains come at the cost of these conventional axes. More importantly, the universal Rubric score---which jointly captures semantic fidelity, visual quality, and rendering style---places \our{} as the top performer, confirming that our approach performs better on the metric that aligns most closely with human judgments in our evaluation.

\input{tables/output_human_study}

\paragraph{Human preference confirms the performance gains.}
To complement the automated evaluation, we conduct a blinded pairwise human preference study on 150 MMSVG-Bench prompts, comparing \our{} with five representative baselines across foundation models, SVG specialists, and diffusion-optimized methods. As shown in Table~\ref{tab:human_preference}, \our{} achieves a non-tie win rate above 50\% against every evaluated baseline, ranging from 53.3\% against VectorFusion to 96.5\% against JanusCoder. These results provide direct human evidence that the improvements of \our{} extend beyond automated metrics. The full annotation protocol is provided in Appendix~\ref{app:human_preference_study}.

\input{tables/rl_comparison}

\subsection{Reward Design Analysis (RQ2)}
\label{sec:rq2}

To isolate the contribution of our reward design, we fix the base model (Qwen3-8B) and the GRPO optimizer, and vary only the reward signal across four configurations. \emph{Zero-Shot} denotes the base model without any RL post-training. \emph{C+A+H RL} optimizes a weighted combination of CLIP, Aesthetic, and HPS scores, representing a typical multi-metric scalar reward. \emph{Universal Rubric RL} replaces this scalar with our universal evaluation rubric, applying an identical, query-agnostic checklist to every sample. \our{} is our full method, which generates a tailored rubric for each query. Results on MMSVG-Illustration and MMSVG-Icon are reported in Table~\ref{tab:rl_comparison}.

\paragraph{\our{} delivers the strongest balanced gains.}
Our full method achieves the highest Rubric scores across both benchmarks, reaching 0.693 on Illustration and 0.683 on Icon, compared with 0.432 and 0.395 for zero-shot. It also improves CLIP, HPS, and Aesthetic over the base model, yielding the strongest overall performance among the evaluated RL reward designs. This indicates that grounding each criterion in the specific query provides a fine-grained and prompt-aligned optimization signal.

\paragraph{Universal rubrics improve steadily but lack instance-level granularity.}
Universal Rubric RL delivers consistent gains over the zero-shot baseline across multiple axes, lifting Rubric to $0.660$ on Illustration and $0.591$ on Icon while keeping CLIP and HPS healthy. This validates the benefit of multi-axis, dimension-decomposed feedback without the severe reward hacking observed with C+A+H RL. However, because the same generic checklist is applied uniformly to every query, the reward signal cannot reflect the prompt-specific notions of correctness that distinguish, \textit{e.g.}, a minimalist icon from a richly detailed illustration. The resulting optimization granularity remains coarser than that of an instance-aware rubric, leaving rubric gaps of $+0.033$ on Illustration and $+0.092$ on Icon relative to our full method.
\paragraph{Conventional metrics collapse into reward hacking.}
C+A+H RL inflates the Aesthetic score to 6.697 on Illustration and 6.210 on Icon—far above all other variants—but degrades other important dimensions: 
CLIP drops from $0.244$ to $0.196$ on Illustration, and Rubric collapses from $0.395$ to $0.262$ on Icon, even falling below the zero-shot baseline. We observe that this hacking is especially severe on Icon: in an attempt to fool the aesthetic classifier, the policy generates densely overlapping strokes and repeated decorative paths, blowing up the average sequence length to $6.3$k tokens (vs.\ $0.3$k for zero-shot) while degrading the actual visual quality. This illustrates a characteristic failure mode of scalar multi-metric rewards: without dimension-aware decomposition, the optimizer concentrates probability mass on whichever signal is easiest to inflate, exactly the pathology that motivates our rubric-based design.

\begin{figure}[t]
\vspace{-10pt}
  \centering
  \includegraphics[width=\columnwidth]{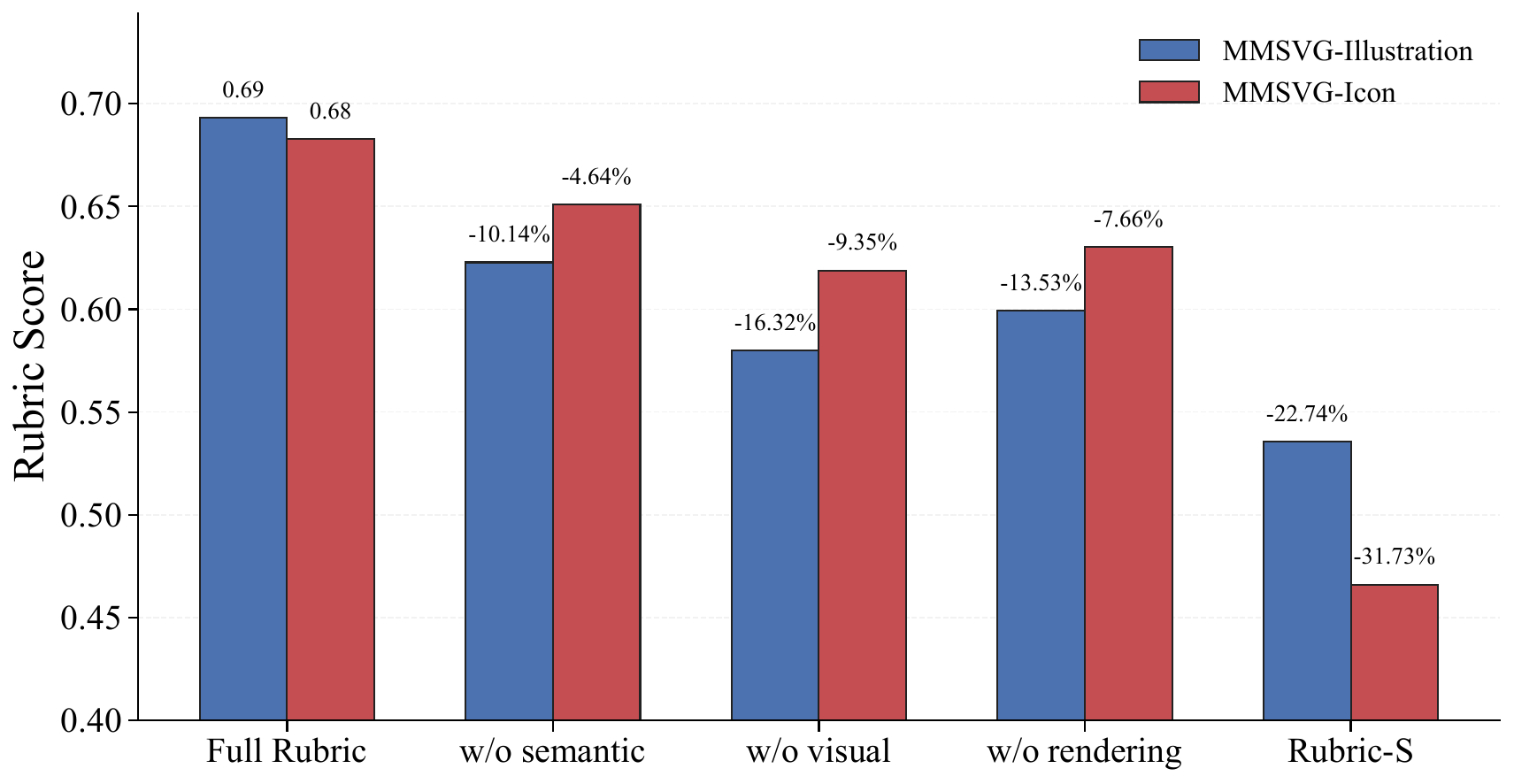}

  \caption{Ablation of rubric design on MMSVG-Illustration and MMSVG-Icon.}

  \label{fig:ablation}
  \vspace{-18pt}
\end{figure}

\subsection{Ablation Study (RQ3)}
\label{sec:ablation}

We ablate two aspects of our rubric design across both MMSVG benchmarks: which evaluation axes drive the final policy, and how sensitive the framework is to the rubric-generation prompt itself. All variants share the same base model (Qwen3-8B), GRPO optimizer, and training hyperparameters. Three variants---\emph{w/o Semantic}, \emph{w/o Visual}, and \emph{w/o Rendering}---zero out the corresponding pair of items (items 1--2, 3--4, 5--6, respectively) during reward aggregation while keeping the remaining items intact. A fourth variant, \emph{Rubric-S}, replaces our default rubric-generation prompt with a stricter, structurally focused alternative that removes the stylistic axis and generates only five rubric items. We further append a fixed text-hint penalty item to this variant, resulting in six scoring items in total. The Rubric-S generation prompt and fixed penalty item are provided in Appendix~\ref{app:rubric_s_prompt}. Rubric scores for all variants are reported in Figure~\ref{fig:ablation}.

\paragraph{Effectiveness of the three rubric axes.}
Removing \emph{Visual Quality} causes the largest decline ($0.693\!\rightarrow\!0.580$), indicating that silhouette, form, and composition items contribute the most non-trivial training signal and are where RL helps the policy most. Removing \emph{Rendering} yields the second-largest drop ($\rightarrow\!0.599$), confirming that craftsmanship items provide complementary diagnostic signal not captured by the other two axes. Removing \emph{Semantic Fidelity} produces the smallest decrease ($\rightarrow\!0.623$)---plausibly because the base Qwen3-8B already exhibits strong text-image alignment from pretraining, so the marginal gain from explicitly rewarding semantic items is smaller than for the visual and stylistic axes that pretraining covers less directly. 
These results show that all axes provide complementary training signals and that the dimensional decomposition is not redundant.

\paragraph{Sensitivity to rubric design.}
Despite its stricter scoring criteria, Rubric-S triggers a previously unobserved \emph{text-hint hacking} behavior and drops the universal Rubric to $0.536$---a larger decline than any single-axis ablation above. 
The policy increasingly embeds readable prompt-related text inside the rendered SVGs (\textit{e.g.}, the literal query word rendered as a stylized label) despite the explicit text-hint penalty, suggesting that under this stricter, structurally focused rubric specification, textual cues remain an easy shortcut for satisfying the resulting criteria.
This finding highlights a subtle property of rubric design: \emph{ complementary axes are at least as important as scoring strictness}, and changing the rubric-generation specification can reopen degenerate optimization channels even when the resulting criteria are more strictly defined.

\input{tables/robustness_base_model}

\input{tables/robustness_rubric_generator}

\begin{figure}[t]
\vspace{-10pt}
  \centering
  \includegraphics[width=\columnwidth]{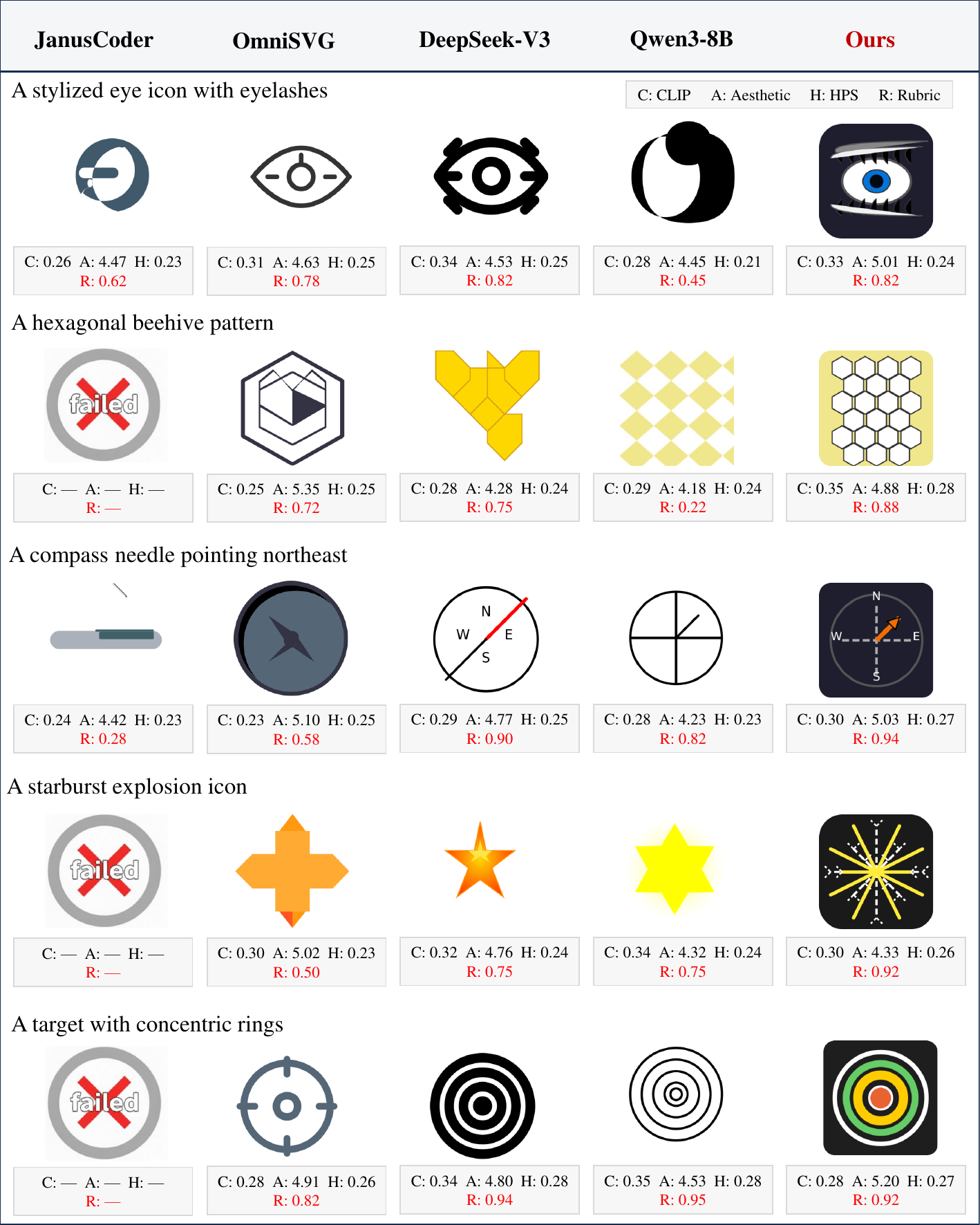}

  \caption{\textbf{Qualitative comparison} of scalable vector graphics generation between \our{} and four baselines.}

  \label{fig:visualization}
  \vspace{-18pt}
\end{figure}

\subsection{Robustness Analysis (RQ4)}
\label{sec:robustness}
To examine whether the effectiveness of \our{} depends on a particular base model or rubric generator, we evaluate the framework across different base-model scales and rubric sources while keeping the remaining training setup unchanged.

\paragraph{Robustness across base models.}
We replace the default Qwen3-8B policy with Qwen3-4B while using the same Claude-Opus-4.6-generated rubrics. As shown in Table~\ref{tab:backbone_transfer}, RULER-4B consistently improves over Qwen3-4B across all metrics, increasing the universal Rubric score from $0.372\pm0.017$ to $0.561\pm0.008$ on Illustration and from $0.338\pm0.017$ to $0.559\pm0.006$ on Icon. RULER-8B further achieves $0.692\pm0.007$ and $0.662\pm0.016$, respectively, remaining ahead of the substantially larger Qwen3-32B under the same evaluation setting ($0.585\pm0.005$ and $0.566\pm0.026$). These results show that the gains of \our{} persist across different base-model scales and remain stable over repeated runs.

\paragraph{Robustness across rubric generators.}
We replace Claude-Opus-4.6 with GPT-5.5~\cite{opus46,gpt55} for rubric generation while keeping the Qwen3-8B policy and the remaining training setup unchanged. As shown in Table~\ref{tab:rubric_generator_main}, both variants consistently improve over the Qwen3-8B base model across all four metrics. In particular, the universal Rubric score reaches $0.662\pm0.016$ with Claude-Opus-4.6 and $0.646\pm0.007$ with GPT-5.5, compared with $0.394\pm0.016$ for the base model. These results indicate that the effectiveness of instance-aware rubric rewards does not depend on a particular rubric generator.

\subsection{Qualitative Analysis (RQ5)}
\label{sec:qualitative}

Figure~\ref{fig:visualization} compares RULER with four baselines on representative icon-style prompts. RULER better preserves prompt-specific details, including eyelashes, honeycomb structures, northeast orientation, starburst shapes, and concentric rings, while producing richer colors and more coherent compositions. Several baselines fail to render or omit key details. These observations are consistent with our instance-aware rubric design, which assesses semantic fidelity, visual quality, and rendering style. Relative to Qwen3-8B, RULER shows clear qualitative gains, with visual quality comparable to DeepSeek-V3 on the illustrated examples.

%% file: tables/main_table.tex
\begin{table*}[t]
    \centering
    \caption{Main results on MMSVG benchmarks. Our method achieves the best universal rubric scores on both benchmarks while maintaining competitive CLIP and HPS performance with significantly fewer tokens than optimization-based methods.}
    \resizebox{\textwidth}{!}{%
    \begin{tabular}{@{} l cccc c cccc c @{}}
        \toprule
        \multirow{2}{*}{\textbf{Method}} 
        & \multicolumn{5}{c}{\textbf{MMSVG-Illustration}} 
        & \multicolumn{5}{c}{\textbf{MMSVG-Icon}} \\
        \cmidrule(lr){2-6} \cmidrule(lr){7-11}
        & CLIP$\uparrow$ & Aesth.$\uparrow$ & HPS$\uparrow$ & Rubric$\uparrow$ & Tokens 
        & CLIP$\uparrow$ & Aesth.$\uparrow$ & HPS$\uparrow$ & Rubric$\uparrow$ & Tokens \\
        \midrule

        \rowcolor{gray!12}
        \multicolumn{11}{c}{\textit{\textbf{Diffusion-Optimized Methods}}} \\
        \midrule
        VectorFusion      & \textbf{0.313}  & \underline{4.930}  & \underline{0.265}  & 0.611  & 31.4k  
                          & \textbf{0.309}  & 4.624  & \textbf{0.254}  & 0.510  & 31.4k  \\
        SVGDreamer        & \underline{0.293}  & \textbf{5.119}  & \textbf{0.266}  & 0.547  & 124.3k 
                          & 0.276  & \underline{4.850}  & \underline{0.253}  & 0.413  & 124.3k \\

        \midrule
        \rowcolor{gray!12}
        \multicolumn{11}{c}{\textit{\textbf{Foundation LLMs}}} \\
        \midrule
        Qwen3-8B          & 0.244  & 4.309  & 0.233  & 0.432  & 0.4k   
                          & 0.266  & 4.424  & 0.234  & 0.395  & 0.3k   \\
        Qwen3-32B         & 0.224  & 4.392  & 0.215  & 0.651  & 0.5k      
                          & 0.228  & 4.514  & 0.218  & \underline{0.681}  & 0.3k      \\
        DeepSeek-V3       & 0.285  & 4.653  & 0.251  & \underline{0.686}  & 0.5k   
                          & \underline{0.290}  & 4.610  & 0.248  & 0.673  & 0.2k   \\

        \midrule
        \rowcolor{gray!12}
        \multicolumn{11}{c}{\textit{\textbf{Specialist Models}}} \\
        \midrule
        IconShop          & 0.236  & 4.480  & 0.219  & 0.262  & 2.6k   
                          & 0.285  & 4.541  & 0.243  & 0.586  & 1.3k   \\
        OmniSVG-8B        & 0.215  & 4.505  & 0.221  & 0.288  & 6.9k   
                          & 0.272  & 4.629  & 0.244  & 0.565  & 5.7k   \\
        JanusCoder-8B     & 0.228  & 4.323  & 0.227  & 0.390  & 1.0k   
                          & 0.256  & 4.432  & 0.234  & 0.384  & 1.0k   \\

        \midrule[0.8pt]
        \textbf{\our{}} 
                          & 0.249  & 4.859  & 0.246  & \textbf{0.693}  & 2.0k   
                          & 0.269  & \textbf{4.873}  & 0.247  & \textbf{0.683}  & 0.7k   \\
        \bottomrule
    \end{tabular}%
    }
    \label{tab:main_results}
\end{table*}

%% file: tables/output_human_study.tex
\begin{table}[t]
\centering
\small
\begin{tabular}{lrrrr}
\toprule
Baseline & Win & Tie & Loss & Win Rate \\
\midrule
Qwen3-8B     & 122 & 14 & 14 & 89.7\% \\
Qwen3-32B    & 84  & 23 & 43 & 66.1\% \\
VectorFusion & 73  & 13 & 64 & 53.3\% \\
OmniSVG      & 93  & 11 & 46 & 66.9\% \\
JanusCoder   & 139 & 6  & 5  & 96.5\% \\
\bottomrule
\end{tabular}
\caption{\textbf{Blinded human preference results.} Win rates compare \our{} against each baseline on 150 MMSVG-Bench prompts and exclude ties.}
\label{tab:human_preference}
\end{table}

%% file: tables/rl_comparison.tex
\begin{table*}[t]
    \centering
    \caption{\textbf{Comparison of different RL reward designs on MMSVG benchmarks.} C, A and H denote CLIP, Aesthetic and HPS, respectively.}
    \label{tab:rl_comparison}
    \resizebox{\textwidth}{!}{%
    \begin{tabular}{@{} l cccc c cccc c @{}}
        \toprule
        \multirow{2}{*}{\textbf{Method}} & \multicolumn{5}{c}{\textbf{MMSVG-Illustration}} & \multicolumn{5}{c}{\textbf{MMSVG-Icon}} \\
        \cmidrule(lr){2-6} \cmidrule(lr){7-11}
        & CLIP$\uparrow$ & Aesth.$\uparrow$ & HPS$\uparrow$ & Rubric$\uparrow$ & Tokens & CLIP$\uparrow$ & Aesth.$\uparrow$ & HPS$\uparrow$ & Rubric$\uparrow$ & Tokens \\
        \midrule
        Zero-Shot            & 0.244  & 4.309  & 0.233  & 0.432  & 0.4k   & 0.266  & 4.424  & 0.234  & 0.395  & 0.3k   \\
        C+A+H RL   & 0.196  & \textbf{6.697}  & 0.235  & 0.464  & 0.7k   & 0.232  & \textbf{6.210}  & 0.239  & 0.262  & 6.3k   \\
        Universal Rubric RL    & 0.244  & 4.813  & 0.241  & 0.660  & 0.7k   & 0.266  & 4.793  & 0.244  & 0.591  & 0.3k   \\
        \textbf{\our{}} & \textbf{0.249}  & 4.859  & \textbf{0.246}  & \textbf{0.693}  & 2.0k   & \textbf{0.269}  & 4.873  & \textbf{0.247}  & \textbf{0.683}  & 0.7k   \\
        \bottomrule
    \end{tabular}%
    }
\end{table*}

%% file: tables/robustness_base_model.tex
\begin{table*}[t]
\centering
\caption{\textbf{Robustness across base-model scales on MMSVG benchmarks.}
We compare RULER-4B and RULER-8B with their corresponding base models and representative baselines. Results are reported as mean $\pm$ standard deviation over five runs.}
\label{tab:backbone_transfer}

\resizebox{\textwidth}{!}{%
\begin{tabular}{@{} l cccc cccc @{}}
\toprule
\multirow{2}{*}{\textbf{Method}}
& \multicolumn{4}{c}{\textbf{MMSVG-Illustration}}
& \multicolumn{4}{c}{\textbf{MMSVG-Icon}} \\
\cmidrule(lr){2-5} \cmidrule(lr){6-9}
& CLIP$\uparrow$
& Aesth.$\uparrow$
& HPS$\uparrow$
& Rubric$\uparrow$
& CLIP$\uparrow$
& Aesth.$\uparrow$
& HPS$\uparrow$
& Rubric$\uparrow$ \\
\midrule

Qwen3-4B
& $0.240{\pm}0.002$
& $4.227{\pm}0.034$
& $0.231{\pm}0.001$
& $0.372{\pm}0.017$
& $0.262{\pm}0.002$
& $4.402{\pm}0.021$
& $0.232{\pm}0.001$
& $0.338{\pm}0.017$ \\

\textbf{RULER-4B}
& $\mathbf{0.244{\pm}0.001}$
& $\mathbf{4.616{\pm}0.020}$
& $\mathbf{0.241{\pm}0.001}$
& $\mathbf{0.561{\pm}0.008}$
& $\mathbf{0.268{\pm}0.001}$
& $\mathbf{4.778{\pm}0.019}$
& $\mathbf{0.244{\pm}0.000}$
& $\mathbf{0.559{\pm}0.006}$ \\

\midrule

Qwen3-8B
& $0.249{\pm}0.003$
& $4.299{\pm}0.015$
& $0.234{\pm}0.001$
& $0.440{\pm}0.011$
& $0.266{\pm}0.003$
& $4.434{\pm}0.015$
& $0.234{\pm}0.001$
& $0.394{\pm}0.016$ \\

Qwen3-32B
& $\mathbf{0.265{\pm}0.004}$
& $4.395{\pm}0.035$
& $0.241{\pm}0.000$
& $0.585{\pm}0.005$
& $\mathbf{0.279{\pm}0.002}$
& $4.483{\pm}0.022$
& $0.242{\pm}0.001$
& $0.566{\pm}0.026$ \\

JanusCoder
& $0.221{\pm}0.018$
& $4.357{\pm}0.053$
& $0.224{\pm}0.006$
& $0.313{\pm}0.109$
& $0.249{\pm}0.006$
& $4.453{\pm}0.031$
& $0.230{\pm}0.003$
& $0.324{\pm}0.040$ \\

\textbf{RULER-8B}
& $0.247{\pm}0.002$
& $\mathbf{4.872{\pm}0.015}$
& $\mathbf{0.246{\pm}0.000}$
& $\mathbf{0.692{\pm}0.007}$
& $0.269{\pm}0.001$
& $\mathbf{4.868{\pm}0.020}$
& $\mathbf{0.247{\pm}0.000}$
& $\mathbf{0.662{\pm}0.016}$ \\

\bottomrule
\end{tabular}%
}
\end{table*}

%% file: tables/robustness_rubric_generator.tex
\begin{table}[t]
\centering
\small
\caption{\textbf{Robustness across rubric generators on MMSVG-Icon.}
RULER uses the same Qwen3-8B policy with different rubric generators. Results are reported as mean $\pm$ standard deviation over five runs.}
\label{tab:rubric_generator_main}

\resizebox{\columnwidth}{!}{%
\begin{tabular}{@{}llcccc@{}}
\toprule
\textbf{Method}
& \textbf{Rubric Generator}
& CLIP$\uparrow$
& Aesth.$\uparrow$
& HPS$\uparrow$
& Rubric$\uparrow$ \\
\midrule

Qwen3-8B
& --
& $0.266{\pm}0.003$
& $4.434{\pm}0.015$
& $0.234{\pm}0.001$
& $0.394{\pm}0.016$ \\

\midrule

\multirow{2}{*}{\textbf{\our{}}}
& Claude-Opus-4.6
& $0.269{\pm}0.001$
& $\mathbf{4.868{\pm}0.020}$
& $\mathbf{0.247{\pm}0.000}$
& $\mathbf{0.662{\pm}0.016}$ \\

& GPT-5.5
& $\mathbf{0.278{\pm}0.001}$
& $4.708{\pm}0.016$
& $0.244{\pm}0.000$
& $0.646{\pm}0.007$ \\

\bottomrule
\end{tabular}%
}
\end{table}

%% file: sections/conclusion.tex
\section{Conclusion}
\label{sec:conclusion}

We first established that evaluating open-ended SVGs with multi-axis rubrics aligns closely with human judgments, achieving strong sample-level correlation and pairwise ranking agreement. Building on this, we introduced RULER, which converts instance-aware rubrics into dense rewards optimized via GRPO. Across MMSVG benchmarks, RULER mitigates reward hacking, outperforms scalar-reward RL baselines and dedicated SVG specialists, and remains consistently effective across different base models and rubric generators.

%% file: sections/limitations.tex
\section*{Limitations}

\our{} improves reward design for open-ended SVG generation, but several limitations remain. First, the framework depends on two external models: a frontier model to generate the instance-aware rubric and a VLM judge to score rendered outputs. As a result, reward quality may inherit their biases, preferences, and failure modes. Although the rubric is more faithful than scalar metrics in our setting, the judge can still overvalue superficial cues or underweight subtle stylistic qualities, especially on prompts that fall outside the training distribution.
Second, our method increases the computational cost of RL. Each training step requires rendering sampled SVGs and querying a judge VLM to score each rendered output against its instance-specific rubric, which is substantially more expensive than using lightweight scalar rewards such as CLIP or heuristic code-based signals. This cost may limit scalability to larger models, longer rollouts, or broader hyperparameter searches.
Third, while instance-aware rubrics preserve open-endedness better than paired-reference objectives, they still impose a particular decomposition of quality into semantic, visual, and stylistic axes. That decomposition is useful in our benchmarks, but it may not fully capture all valid artistic intents or domain-specific preferences. Extending rubric design to interactive, human-steerable, or domain-adaptive settings remains future work.

%% file: sections/acknowledgement.tex
\section*{Acknowledgments} 
This work was supported by the Ant Group Research Intern Program.
We also sincerely thank Zhaoyang Zhang, Cong Chen, and Hailong Sun
for their valuable discussions, support, and helpful feedback.

%% file: sections/appendix.tex
\appendix

\section{Training Details}
\label{app:training_details}

We train the policy with Group Relative Policy Optimization (GRPO) using the VERL framework and the vLLM rollout engine. Unless otherwise specified, the policy backbone is Qwen3-8B, and training uses the prompt--rubric pairs described in Appendix~\ref{app:dataset_construction}. We disable thinking mode in the chat template, set the maximum prompt length to 512 tokens and the maximum response length to 4096 tokens, and sample eight rollouts per prompt with a temperature of 1.0. We use a training batch size of 128, a PPO mini-batch size of 64, and a PPO micro-batch size of 8 per GPU. The learning rate is $1\times10^{-6}$, and the entropy coefficient is 0.001. We do not use KL regularization in either the reward or the actor loss.

The rollout model uses bfloat16 precision, tensor parallelism of 8, gradient checkpointing, and remove-padding optimization. The maximum token budget per GPU is 65,536 for log-probability computation. All experiments are conducted on one node with 8 H800 80~GB GPUs.

\section{Dataset Construction}
\label{app:dataset_construction}

We construct the RL training data from the training splits of MMSVG-Icon and MMSVG-Illustration. We randomly sample 20K prompts from MMSVG-Icon and 12K prompts from MMSVG-Illustration. For each prompt, we generate an instance-aware rubric using the template in Appendix~\ref{app:rubric_generation_prompt}. The generated rubric is paired with its source prompt and serves as the reward specification during RL training.

We apply both structural and quality filtering to the generated rubrics. We first discard cases in which rubric generation fails or the returned rubric does not satisfy the required format or item structure. We then perform a rubric-quality check using the ideal SVG generated together with each rubric. Specifically, the ideal SVG is rendered and evaluated by Qwen3-VL-8B~\cite{qwen3vl} against its corresponding rubric using the same item-level scoring and normalized weighted aggregation used during RL reward computation. Rubrics whose ideal SVG receives a reward below 0.9 are removed. This step filters out rubrics that are poorly aligned with the ideal SVG used during rubric construction.

After filtering, the final training sets contain 19,531 prompt--rubric pairs for MMSVG-Icon and 11,206 for MMSVG-Illustration, for a total of 30,737 training examples.

The MMSVG training and test prompts are generated separately, reducing the risk of direct prompt-level overlap between RL training and evaluation. Because both splits remain within the same benchmark distribution, however, we do not treat this setting as a domain-shifted out-of-distribution evaluation.

\section{Reward Computation and SVG Rendering}
\label{app:reward_rendering}

The reward is produced by a rubric-based VLM judge. We use Qwen3-VL-8B as the frozen judge and serve it through multiple OpenAI-compatible endpoints for parallel evaluation. For each generated SVG, we first render the code into an image and then provide the judge with the original text prompt, the rendered candidate, and the corresponding instance-aware rubric. The judge scores the six rubric items independently, producing item-level satisfaction scores $s_i\in[0,1]$. The final reward is the normalized weighted average
\[
R = \frac{\sum_{i=1}^{6} w_i s_i}{\sum_{i=1}^{6} w_i},
\]
where $w_i$ denotes the weight of the $i$-th rubric item.

For each prompt, the policy samples $G=8$ rollouts. Every rollout is rendered with CairoSVG and scored independently by Qwen3-VL-8B. The resulting rewards are normalized within the rollout group and used to form the relative advantages for the GRPO update. The judge remains fixed throughout training. When comparing reward designs, the policy backbone, optimizer, and other training settings are held fixed; the Qwen3-VL-8B judge is also fixed across the rubric-based reward variants.

The per-endpoint concurrency is 32, and the judge timeout is 180 seconds. Blank renderings are detected by comparing the rendered image with a white image using a mean-squared-error threshold of $10^{-4}$.

The standard RULER reward consists only of the six instance-aware rubric items and does \emph{not} include an auxiliary text-hint penalty. Readable-text shortcuts are discouraged within the rubric itself by requiring the prompt concept to be communicated visually and by penalizing text shortcuts under the rendering-quality criterion. In contrast, the Rubric-S ablation uses a different five-item rubric together with an additional fixed text-hint penalty item with weight $-2$, as detailed in Appendix~\ref{app:rubric_s_analysis}.

All generated SVGs are rendered with CairoSVG 2.9.0 at $512\times512$ resolution on a white background before reward evaluation and metric computation. Automated metrics are aggregated over successfully rendered samples. We use the same rendering pipeline for RULER and all baselines and report model-specific render success rates in Appendix~\ref{app:render_success}.

\section{Human Evaluation}
\label{app:human_evaluation}

We conduct two human studies with complementary purposes. The first evaluates the alignment between automated metrics and human judgments, while the second directly compares the final outputs of RULER with representative baselines.

\subsection{Human Alignment Study}
\label{app:human_alignment_protocol}

To assess how well different automated metrics reflect human judgments of SVG quality, we sample 300 prompts, evenly divided between MMSVG-Icon and MMSVG-Illustration. For each prompt, we collect one output from Qwen3-8B, Qwen3-32B, and Claude-Opus-4.6, yielding 900 rendered SVGs in total.

Ten human annotators participate in the annotation. The rendered outputs are anonymized and randomly assigned to annotators. Each sample is assigned a single holistic quality score from 0 to 100, with annotators considering prompt fidelity, visual quality, composition, rendering quality, and stylistic coherence when making the judgment. After the initial annotation, three validators review all assigned scores and rescore cases judged to be unreasonable.

For the automated rubric-based evaluation in this study, we use GPT-5-mini with a shared universal rubric. This evaluator is separate from the reward model used during RULER training: training uses Qwen3-VL-8B to score prompt-specific instance-aware rubrics, whereas the human-alignment analysis uses GPT-5-mini with the same universal evaluation rubric for all samples.

Using the human scores as the reference, the universal Rubric score achieves a Spearman rank correlation of $\rho=0.7929$, compared with $0.6051$ for Aesthetic and $0.5518$ for CLIP. For pairwise ranking agreement, the corresponding Goodman--Kruskal Gamma is $\gamma=0.7574$, compared with $0.5465$ for Aesthetic and $0.5295$ for CLIP.

\subsection{Final-Output Human Preference Study}
\label{app:human_preference_study}

We further conduct a blinded pairwise preference study to directly evaluate the final outputs of RULER. We sample 150 prompts from MMSVG-Bench and compare the RULER output for each prompt with the corresponding output from five representative baselines: Qwen3-8B, Qwen3-32B, VectorFusion, OmniSVG, and JanusCoder. This produces 750 pairwise comparisons in total.

Three human annotators participate in the study. The 750 comparisons are randomly assigned among the evaluators, with each pair evaluated once. For each comparison, the evaluator is shown only the text prompt and two rendered SVG candidates. Model identities are hidden, and the left--right order of the candidates is randomized. The evaluator selects whether RULER wins, ties, or loses based on prompt fidelity and overall visual quality. If one candidate fails to render while the other renders successfully, the rendering failure is counted as a loss.

We report the non-tie win rate as
\[
\mathrm{WinRate}
=
\frac{\mathrm{Win}}
{\mathrm{Win}+\mathrm{Loss}},
\]
with ties excluded from the denominator.

\begin{table}[t]
\centering
\small
\begin{tabular}{lrrrr}
\toprule
Comparison & Win & Tie & Loss & Win Rate \\
\midrule
RULER vs.\ Qwen3-8B     & 122 & 14 & 14 & 89.7\% \\
RULER vs.\ Qwen3-32B    & 84  & 23 & 43 & 66.1\% \\
RULER vs.\ VectorFusion & 73  & 13 & 64 & 53.3\% \\
RULER vs.\ OmniSVG      & 93  & 11 & 46 & 66.9\% \\
RULER vs.\ JanusCoder   & 139 & 6  & 5  & 96.5\% \\
\bottomrule
\end{tabular}
\caption{Blinded pairwise human preference results on 150 MMSVG-Bench prompts for each baseline comparison. Win rates are computed after excluding ties.}
\label{tab:human_preference_appendix}
\end{table}

RULER achieves a non-tie win rate above 50\% against all five evaluated baselines.

\section{Additional Evaluation and Robustness Analyses}
\label{app:additional_analysis}

\subsection{Stability Across Inference Runs}
\label{app:five_run_eval}

The main benchmark results are obtained from a single inference run. To quantify sensitivity to decoding randomness, we additionally evaluate RULER and representative autoregressive baselines over five independently seeded inference runs. The resulting mean and standard deviation are reported in Table~\ref{tab:backbone_transfer}.

Across the five runs, RULER achieves a Rubric score of $0.692\pm0.007$ on MMSVG-Illustration and $0.662\pm0.016$ on MMSVG-Icon, outperforming both its Qwen3-8B backbone and the larger Qwen3-32B model on this metric. The small run-to-run variation further shows that the improvement is stable across decoding seeds.

\subsection{Render Success Rates}
\label{app:render_success}

Because automated metrics are computed over successfully rendered outputs, we separately report render success rates for all methods in Table~\ref{tab:render_success}. Under the single-run setting used for the main benchmark comparison, each method generates one output for each of 150 MMSVG-Icon prompts and 150 MMSVG-Illustration prompts.

\begin{table}[t]
\centering
\small
\begin{tabular}{lcc}
\toprule
Method & MMSVG-Icon & MMSVG-Illustration \\
\midrule
RULER        & 99.3\%  & 100.0\% \\
OmniSVG      & 99.3\%  & 99.3\% \\
DeepSeek-V3  & 100.0\% & 99.3\% \\
JanusCoder   & 68.0\%  & 68.7\% \\
Qwen3-8B     & 94.7\%  & 89.3\% \\
Qwen3-32B    & 99.3\%  & 100.0\% \\
IconShop     & 100.0\% & 100.0\% \\
VectorFusion & 100.0\% & 100.0\% \\
SVGDreamer   & 100.0\% & 100.0\% \\
\bottomrule
\end{tabular}
\caption{Render success rates under the single-run benchmark evaluation.}
\label{tab:render_success}
\end{table}

RULER successfully renders 99.3\% of Icon samples and all Illustration samples. Thus, its aggregate metric values are affected by very few excluded samples, while the reported success rates provide additional context for methods with more frequent rendering failures.

\subsection{Policy-Scale Analysis}
\label{app:smaller_backbone}

% We further instantiate RULER with a smaller Qwen3-4B policy to examine the effect of policy scale. Table~\ref{tab:backbone_transfer} extends the five-run comparison in Table~\ref{tab:five_run_eval} by adding Qwen3-4B and its RULER-trained counterpart. To distinguish the two policy scales in this analysis, we denote the standard Qwen3-8B-based model as RULER-8B and the smaller variant as RULER-4B. All results are averaged over five independently seeded inference runs.

We further instantiate RULER with a smaller Qwen3-4B policy to examine the effect of policy scale. We compare RULER-4B and RULER-8B with their corresponding base models and representative baselines in Table~\ref{tab:backbone_transfer}. To distinguish the two policy scales in this analysis, we denote the standard Qwen3-8B-based model as RULER-8B and the smaller variant as RULER-4B. All results are averaged over five independently seeded inference runs.

RULER improves both policy backbones substantially on the Rubric metric. For Qwen3-4B, the score increases from $0.372$ to $0.561$ on Illustration and from $0.338$ to $0.559$ on Icon; for Qwen3-8B, it increases from $0.440$ to $0.692$ and from $0.394$ to $0.662$, respectively. RULER-4B also approaches the much larger Qwen3-32B on Rubric ($0.561$ vs. $0.585$ on Illustration and $0.559$ vs. $0.566$ on Icon), while RULER-8B exceeds it on both benchmarks. These results show that the gains from instance-aware rubric rewards are retained when the policy is scaled down from 8B to 4B.

\subsection{Rubric Generator Analysis}
\label{app:rubric_generator_transfer}

The default training rubrics are generated with Claude-Opus-4.6. We additionally generate a separate rubric set with GPT-5.5 for MMSVG-Icon and retrain the same Qwen3-8B policy under otherwise identical settings. Table~\ref{tab:rubric_generator_transfer_appendix} places the two RULER variants alongside the same five-run autoregressive baselines used above. Both RULER rows use the Qwen3-8B policy; only the model used to generate the training rubrics is changed.

\begin{table*}[t]
\centering
\caption{\textbf{Robustness to different rubric generators on MMSVG-Icon.} Results are averaged over five independently seeded inference runs and reported as mean $\pm$ standard deviation.}
\label{tab:rubric_generator_transfer_appendix}

\resizebox{\textwidth}{!}{%
\begin{tabular}{@{} l l cccc @{}}
\toprule
\textbf{Method}
& \textbf{Rubric Generator}
& CLIP$\uparrow$
& Aesth.$\uparrow$
& HPS$\uparrow$
& Rubric$\uparrow$ \\
\midrule

Qwen3-8B
& -- 
& $0.266{\pm}0.003$
& $4.434{\pm}0.015$
& $0.234{\pm}0.001$
& $0.394{\pm}0.016$ \\

Qwen3-32B
& --
& $\mathbf{0.279{\pm}0.002}$
& $4.483{\pm}0.022$
& $0.242{\pm}0.001$
& $0.566{\pm}0.026$ \\

JanusCoder
& --
& $0.249{\pm}0.006$
& $4.453{\pm}0.031$
& $0.230{\pm}0.003$
& $0.324{\pm}0.040$ \\

\midrule

\multirow{2}{*}{\textbf{\our{}}}
& Claude-Opus-4.6
& $0.269{\pm}0.001$
& $\mathbf{4.868{\pm}0.020}$
& $\mathbf{0.247{\pm}0.000}$
& $\mathbf{0.662{\pm}0.016}$ \\

& GPT-5.5
& $0.278{\pm}0.001$
& $4.708{\pm}0.016$
& $0.244{\pm}0.000$
& $0.646{\pm}0.007$ \\

\bottomrule
\end{tabular}%
}
\end{table*}

Both rubric generators yield RULER models that substantially outperform the Qwen3-8B backbone on Rubric ($0.394$) and also exceed Qwen3-32B ($0.566$). The GPT-5.5-generated rubrics achieve a higher CLIP score, whereas the Claude-Opus-4.6-generated rubrics perform better on Aesthetic, HPS, and Rubric ($0.662$ vs. $0.646$). The comparable performance of the two variants indicates that RULER is not tied to a single rubric generator.

\subsection{Comparison with Iterative Diffusion Methods}
\label{app:diffusion_efficiency}

RULER and diffusion-optimized SVG methods use fundamentally different inference procedures. VectorFusion performs iterative optimization separately for each input prompt, requiring an average of 72.1 minutes per sample and producing SVGs with an average length of 31.4K tokens in our evaluation. In contrast, RULER generates SVG code end-to-end in a single autoregressive pass.

Despite avoiding prompt-specific iterative optimization, RULER achieves higher universal Rubric scores on both benchmarks: $0.683$ versus $0.510$ on MMSVG-Icon and $0.693$ versus $0.611$ on MMSVG-Illustration. We report this comparison to complement the conventional CLIP, Aesthetic, and HPS metrics, on which diffusion-optimized approaches can remain competitive or stronger.

\section{Rubric Design Details}
\label{app:rubric_design_analysis}

\subsection{Ground-Truth-Free Rubric Construction}
\label{app:rubric_construction}

In settings with a reference answer, rubric generation can condition on both the input and the answer. Open-ended text-to-SVG generation, however, does not provide a unique ground-truth SVG. RULER therefore constructs each rubric without a paired reference SVG; the only external task input to the rubric generator is the text instruction.

As specified in Appendix~\ref{app:rubric_generation_prompt}, the generator first produces one ideal SVG as a plausible high-quality realization of the instruction and then derives an instance-specific rubric for evaluating other candidates. The ideal SVG serves as an intermediate quality target during rubric construction rather than as a unique ground-truth answer. The generation prompt explicitly requires the resulting rubric to remain open-ended and prohibits exact reconstruction requirements on geometry, placement, colors, part counts, or other implementation details unless they are specified by the instruction. The ideal SVG is not used as a reference when scoring policy rollouts during RL.

This procedure allows the rubric to capture prompt-specific visual requirements without paired ground-truth SVGs while preserving the one-to-many nature of text-to-SVG generation. For consistency across prompts, we use a fixed six-item structure spanning Semantic Fidelity, Visual Quality, and Rendering Style, with weights $(5,5,5,4,5,5)$. The structure and weights are shared across all prompts, while item titles, descriptions, and scoring guides are generated separately for each instruction.

\subsection{Rubric-S}
\label{app:rubric_s_analysis}

Rubric-S uses an alternative rubric-generation prompt with a stronger emphasis on structural completeness and stricter score calibration. It generates five items---Overall Prompt Readability, Major Component Legibility, Subject Shape Complexity, Key Internal Detail Coverage, and Visual Layering---and removes the stylistic axis used by the default rubric. During training, we additionally append a fixed text-hint penalty item with weight $-2$, resulting in six scoring items in total. A higher satisfaction score on this item indicates stronger evidence of a textual shortcut and therefore reduces the overall reward. For Rubric-S, this negative-weight penalty contributes only to the numerator, while the normalization denominator is computed from the five positive rubric-item weights. The complete prompt is provided in Appendix~\ref{app:rubric_s_prompt}.

Despite the additional text-hint penalty, Rubric-S obtains a universal Rubric score of 0.536 on MMSVG-Illustration, below the full RULER setting and all three single-axis ablations. We also observe that the resulting policy increasingly inserts readable prompt-related text into the rendered SVGs. This result illustrates that changes to the rubric specification can substantially affect optimization behavior, and that stricter scoring criteria do not necessarily produce a better reward signal.

\section{Rubric Generation Cost}
\label{app:rubric_cost}

Instance-aware rubrics are generated once as an offline preprocessing step and cached for subsequent RL training. Table~\ref{tab:rubric_cost} summarizes the cost of generating the rubrics used to construct our RL training data. Claude-Opus-4.6 processes 60.18M input tokens and generates 68.49M output tokens, with a total cost of \$2,013.12. The average rubric-generation cost is \$0.0655 per prompt.

\begin{table}[!ht]
\centering
\footnotesize
\setlength{\tabcolsep}{2.5pt}

\resizebox{\columnwidth}{!}{%
\begin{tabular}{@{}lrrr@{}}
\toprule
Statistic & MMSVG-Icon & MMSVG-Illustration & Overall \\
\midrule
Prompts & 19,531 & 11,206 & 30,737 \\
Avg. Input Tokens & 1,959 & 1,957 & 1,958 \\
Avg. Rubric Tokens & 1,010 & 1,057 & 1,027 \\
Avg. Output Tokens & 1,934 & 2,740 & 2,228 \\
Total Cost & \$1,135.77 & \$877.35 & \$2,013.12 \\
Cost / Prompt & \$0.0582 & \$0.0783 & \$0.0655 \\
\bottomrule
\end{tabular}%
}

\caption{Rubric-generation cost for the RL training data. Avg. Output Tokens includes both the generated ideal SVG and rubric; Avg. Rubric Tokens counts the rubric portion only.}
\label{tab:rubric_cost}
\end{table}

The rubric generator is not queried during policy optimization. The complete rubric-generation and judging prompts are provided in Appendix~\ref{app:prompt_templates}. We will release the generated rubrics together with the processed training data and code.

\section{Prompt Templates}
\label{app:prompt_templates}

\subsection{Rubric Generation Prompt}
\label{app:rubric_generation_prompt}

\begin{prompt}{Rubric Generation for Text-to-SVG Evaluation}
You are an expert SVG designer and an expert rubric writer for open-ended text-to-SVG evaluation. \\[3pt]

\textbf{Task.}
Your task has two sequential stages: \\[2pt]
1. Create one ideal SVG for the given prompt. \\
2. Generate an aligned, instance-specific rubric for judging other rendered SVG candidates for the same prompt. \\[3pt]

The ideal SVG should establish a strong visual reference. The rubric should convert that reference into open-ended evaluation criteria for reinforcement learning. \\[5pt]

\textbf{Core Goal.}
The goal is not merely to make the prompt recognizable. The goal is to encourage polished, deliberate, visually refined SVG artwork. \\[3pt]

Avoid crude or placeholder-like drawings, generic primitive-symbol assemblies, rough sketches, accidental ugliness, visually empty outputs, and readable text labels used as shortcuts. Simplicity is allowed, but only when it looks intentional, balanced, and aesthetically designed. \\[5pt]

\textbf{Stage 1: Ideal SVG Generation.}
First, internally imagine a strong visual solution for the prompt and then write one ideal SVG. The SVG must clearly communicate the prompt concept without using readable text labels. It should look like polished vector artwork, with clear design intention, clean contours, balanced proportions, coherent graphic treatment, and enough visual refinement to serve as a high-quality aesthetic exemplar. It should not be unnecessarily complex or cluttered. This SVG is only one possible good solution and is not the only correct answer. \\[5pt]

\textbf{Stage 2: Rubric Generation.}
Generate a rubric for evaluating other rendered SVG candidates for the same prompt. The rubric will be used as a reinforcement-learning reward signal. It must reward candidates that are semantically correct, visually refined, deliberately composed, and aesthetically stronger than crude recognizable placeholders. \\[3pt]

The rubric must remain open-ended: a candidate may score well with different shapes, layout, orientation, colors, or implementation details. Do not require exact reproduction of the ideal SVG. Do not turn the rubric into a reconstruction checklist. Do not require exact local details, exact geometry, exact placement, exact part counts, or exact colors unless explicitly required by the prompt. \\[5pt]

\textbf{Instance-specific Requirements.}
Items 1--2 must specify the visible semantic evidence needed for this prompt, including concept identity, major components, relationships, states, or distinctive cues when relevant. Items 3--6 must specify the visual form, composition, rendering, and style qualities expected for this prompt. Avoid criteria that could apply unchanged to any SVG prompt. Each item should identify visible qualities that meaningfully distinguish stronger and weaker candidates for this prompt. \\[5pt]

\textbf{Orthogonality Requirements.}
The rubric must contain exactly six items in the fixed order below. Each item should target one distinct observable dimension, be independently judgeable from visible evidence, and penalize a different type of failure. If two items could be justified using the same evidence, rewrite them until they are clearly separable. Do not include catch-all items such as overall beauty, overall quality, overall polish, or overall visual appeal. Do not combine multiple axes in one item. \\[5pt]

\textbf{Fixed Rubric Items and Weights.} \\[2pt]
1. Concept Identity and Overall Readability -- weight 5 \\
2. Major Components and Distinctive Cues -- weight 5 \\
3. Silhouette and Form Refinement -- weight 5 \\
4. Composition and Canvas Design -- weight 4 \\
5. Rendering Finish and Visual Craftsmanship -- weight 5 \\
6. Style Cohesion and Designed Visual Interest -- weight 5 \\[5pt]

\textbf{Item Definitions.} \\[2pt]
\textbf{1. Concept Identity and Overall Readability:}
Judge whether the overall prompt meaning and primary concept are visually understandable without relying on readable text labels. Do not turn this item into a checklist of parts, and do not discuss component richness, visual complexity, composition, rendering polish, or style quality. \\[2pt]

\textbf{2. Major Components and Distinctive Cues:}
Judge whether the major objects, important parts, actions, states, relationships, or concept-distinguishing visual cues are clearly readable. Do not discuss shape elegance, composition quality, rendering cleanliness, visual polish, or style cohesion. \\[2pt]

\textbf{3. Silhouette and Form Refinement:}
Judge the quality of the subject's main shapes, contours, proportions, and silhouette. Distinguish refined, intentional forms from crude blobs, awkward primitive shapes, stiff assembly, collapsed geometry, or weak silhouette design. Do not discuss composition, rendering artifacts, or style consistency. \\[2pt]

\textbf{4. Composition and Canvas Design:}
Judge subject scale, placement, spacing, cropping, negative space, visual balance, and spatial hierarchy. Distinguish deliberately composed images from cramped, tiny, badly cropped, awkwardly placed, or spatially careless ones. Do not discuss shape quality, rendering cleanliness, or style consistency. \\[2pt]

\textbf{5. Rendering Finish and Visual Craftsmanship:}
Judge execution quality and completion. Distinguish finished vector artwork from outputs with broken contours, disconnected parts, stray fragments, noisy marks, malformed geometry, unfinished regions, text shortcuts, or obvious rendering artifacts. Do not discuss composition, shape elegance, or style harmony. \\[2pt]

\textbf{6. Style Cohesion and Designed Visual Interest:}
Judge whether the image has a coherent vector-art treatment and enough intentional visual interest for this prompt. Consider stroke/fill treatment, detail density, color or tone handling, internal structure, decorative accents, depth cues, or graphic rhythm when relevant. Do not discuss semantic correctness, composition, or basic rendering cleanliness. \\[5pt]

\textbf{Scoring Guide Requirements.}
Each rubric item must include a \texttt{scoring\_guide} field. The scoring guide must describe how to assign continuous satisfaction scores from 0.0 to 1.0, use score ranges rather than fixed anchor scores, explicitly allow any decimal value within the ranges, be specific to the current prompt and item, and define visible differences between excellent, strong, partial, weak, and absent satisfaction. \\[3pt]

Use this exact range structure: \\[2pt]
\texttt{0.9-1.0: ...; 0.7-0.8: ...; 0.4-0.6: ...; 0.1-0.3: ...; 0.0: ...} \\[3pt]

Reserve 0.9--1.0 for excellent, visually convincing satisfaction with clear item-specific evidence. Use 0.7--0.8 only for clearly strong satisfaction. Use 0.4--0.6 for partial but visibly flawed satisfaction. Use 0.1--0.3 for weak, ambiguous, fragmentary, primitive, or barely visible evidence. Use 0.0 for absent, contradicted, unreadable, or completely failed evidence. \\[5pt]

\textbf{Writing Rules.}
Return exactly six items in the fixed order above. Each item must have a short, natural, sample-specific title. Each title should mention the prompt object or concept when natural. Each description must be exactly one sentence, self-contained, and positively state the high-quality standard. Do not write generic empty phrases such as ``looks good'', ``well-formed'', or ``coherent style'' without explaining what that means for this instance. \\[5pt]

\textbf{Output Format.}
Return only two Markdown code blocks in this exact order: first, the ideal SVG in an XML code block; second, the rubric items in a JSON code block. \\[3pt]

\textbf{Input prompt:} \texttt{\{description\}}
\end{prompt}

\subsection{Rubric-based Judge Prompt}
\label{app:judge_prompt}

\begin{prompt}{Rubric-based VLM Judging}
You are an expert visual evaluator for a text-to-SVG generation task. You will evaluate a rendered image produced from SVG code. \\[5pt]

\textbf{Inputs.}
You are given the user prompt, the candidate image, and a rubric with several evaluation items. \\[3pt]

\textbf{User prompt:} \texttt{\{user\_prompt\}} \\[3pt]

\textbf{Rubric:} \texttt{\{rubric\_items\_formatted\}} \\[5pt]

Evaluate the candidate image separately for each rubric item. \\[5pt]

\textbf{Evaluation Rules.} \\[2pt]

\textbf{1. Strictly Follow the Rubric.}
The rubric is the complete evaluation standard. Do not introduce requirements that are not stated in the current rubric item. Base your judgment only on visible evidence in the image. Do not guess intent. \\[3pt]

\textbf{2. Continuous Band Scoring.}
Each rubric item includes a scoring guide defining continuous score bands, such as 0.9--1.0 or 0.4--0.6. First, map the visible evidence to the correct score band. Then, assign a precise decimal value that reflects the exact execution quality within that tier. Do not default to high scores just because the image is not blank or broadly recognizable. \\[3pt]

\textbf{3. Score Calibration.}
Scores of 0.9--1.0 are reserved for near-perfect satisfaction with no visible weakness. Scores of 0.7--0.8 mean strong but not exceptional. Scores of 0.45--0.6 indicate mediocre satisfaction: the image partially satisfies the item but has obvious limitations. Scores of 0.2--0.4 indicate weak or flawed evidence. Scores of 0.0--0.15 indicate absent, contradicted, or failed evidence. Do not inflate scores. ``Not blank'' and ``roughly recognizable'' do not justify scores above 0.6. \\[3pt]

\textbf{4. Independent Scoring.}
Score each rubric item independently. Do not let the score of one item affect the score of any other item. \\[5pt]

\textbf{Output Format.}
Return only a valid JSON object. Write the \texttt{reason} before the \texttt{satisfaction} score. \\[3pt]

\texttt{\{} \\
\quad \texttt{"items": [} \\
\quad\quad \texttt{\{} \\
\quad\quad\quad \texttt{"item\_index": 1,} \\
\quad\quad\quad \texttt{"title": "Item Title",} \\
\quad\quad\quad \texttt{"reason": "Briefly evaluate the visible evidence against the scoring guide.",} \\
\quad\quad\quad \texttt{"satisfaction": 0.0} \\
\quad\quad \texttt{\}} \\
\quad \texttt{]} \\
\texttt{\}}
\end{prompt}

\subsection{Universal Rubric Judge Prompt}
\label{app:universal_judge_prompt}

\begin{prompt}{Universal Rubric VLM Judging}
You are an expert visual evaluator for open-ended text-to-SVG generation. \\[3pt]

You will evaluate a rendered image produced from SVG code. \\[5pt]

\textbf{Inputs.}
You are given the user prompt and the candidate image. \\[3pt]

\textbf{User prompt:} \texttt{\{user\_prompt\}} \\[5pt]

Your task is to assign a \textbf{single overall satisfaction score} from 0.0 to 1.0 based on a comprehensive evaluation of the image. \\[5pt]

\textbf{The 6 Evaluation Dimensions.} \\[3pt]

To determine the overall score, evaluate the image across the following six dimensions. The goal is to reward polished, deliberate, visually refined SVG artwork and penalize crude primitive assemblies or rendering artifacts. \\[3pt]

\textbf{1. Concept Identity:}
Is the overall prompt meaning and primary concept visually understandable without relying on readable text labels? \\[2pt]

\textbf{2. Major Components:}
Are the specific major objects, important parts, states, or distinctive visual cues required by the prompt clearly present? \\[2pt]

\textbf{3. Silhouette and Form Refinement:}
Are the main shapes, contours, and proportions refined and intentional? Penalize crude blobs or awkward primitive shape assemblies. \\[2pt]

\textbf{4. Composition and Canvas Design:}
Is the subject properly scaled, well-placed, and spatially balanced within the canvas? Penalize cramped, badly cropped, or spatially careless layouts. \\[2pt]

\textbf{5. Rendering Finish and Visual Craftsmanship:}
Is the vector execution clean and finished? Penalize broken contours, disconnected parts, stray fragments, or visual noise. \\[2pt]

\textbf{6. Style Cohesion and Designed Visual Interest:}
Does the image have a coherent vector-art aesthetic and intentional design? Penalize visually empty outputs, generic placeholders, or inconsistent graphic treatments. \\[5pt]

\textbf{Holistic Scoring Guide.} \\[3pt]

Based on the six dimensions above, assign the final continuous decimal score using the following strict calibration guide. \\[3pt]

\textbf{0.9--1.0 (Excellent \& Refined):}
Outstanding across all six dimensions. The concept is instantly readable, forms are beautifully refined, rendering is flawless, and the aesthetic is highly cohesive. This range is reserved only for convincing, high-quality vector artwork. \\[3pt]

\textbf{0.7--0.8 (Strong but not exceptional):}
Strong performance across most dimensions. The concept is clear (Dims 1--2) and execution is solid (Dims 3--5). The image may lack exceptional stylistic flair (Dim 6) or contain very minor imperfections in form or composition. \\[3pt]

\textbf{0.4--0.6 (Mediocre / Baseline):}
The image partially satisfies the concept (Dims 1--2) but suffers from obvious flaws in Dims 3--6. Typical examples include crude primitive shapes, awkward spatial layouts, lack of finish, or a highly generic ``clipart'' feel. If an image is merely ``not blank'' and ``roughly recognizable'', it belongs in this range. Do not inflate scores above 0.65 for basic, unrefined shape assemblies. \\[3pt]

\textbf{0.1--0.3 (Weak / Flawed):}
The image fails significantly across multiple dimensions. It barely hints at the concept or suffers from severe issues such as collapsed geometry, completely broken rendering, or extreme spatial imbalance. \\[3pt]

\textbf{0.0 (Failed):}
The image is completely unrecognizable, unrelated to the prompt, or visual garbage. \\[5pt]

\textbf{Evaluation Rules.} \\[2pt]

\textbf{1. Continuous Scoring.}
Use any precise decimal value within the score bands, such as 0.45 or 0.72. \\[3pt]

\textbf{2. Anti-Inflation.}
Do not default to high scores merely because the model successfully rendered something. \\[3pt]

\textbf{3. No Text Shortcuts.}
If the image relies on readable text labels to communicate the prompt instead of drawing the concept, severely penalize the overall score. \\[5pt]

\textbf{Output Format.}
Return only a valid JSON object. Write the \texttt{reason} before the \texttt{overall\_score}. In the reason, explicitly comment on how the image performs across the six dimensions---Concept, Components, Form, Composition, Rendering, and Style---before concluding the final score. \\[3pt]

\texttt{\{} \\
\quad \texttt{"reason": "Step-by-step evaluation of the 6 dimensions, followed by the justification for the final score band.",} \\
\quad \texttt{"overall\_score": 0.0} \\
\texttt{\}}
\end{prompt}

\subsection{Rubric-S Generation Prompt}
\label{app:rubric_s_prompt}

\begin{prompt}{Rubric-S: Structural Rubric Generation}
You are an expert SVG designer and an expert rubric writer for open-ended text-to-SVG evaluation. \\[3pt]

\textbf{Task.}
Your task has two sequential stages: \\[2pt]
1. Create one ideal SVG for the given query. \\
2. Generate an aligned evaluation rubric for judging other rendered SVG candidates for the same query. \\[3pt]

The ideal SVG is used only to establish a strong quality target. The rubric must translate that target into an open-ended scoring standard for rollout candidates. Different valid SVG implementations should be able to score highly, while the rubric should be strict enough to avoid inflated base scores and create fine-grained differences between weak, mediocre, good, and excellent rollout candidates. \\[5pt]

\textbf{Core Goal.}
The user query may be simple, but the ideal SVG should not be crude, placeholder-like, or low-effort. Instead, it should be semantically aligned with the query, visually readable, richer than a primitive placeholder, detailed enough to express the subject clearly, structured enough to avoid a flat or underdeveloped appearance, and appropriate as polished vector artwork. Simplicity is allowed only if it is deliberate and visually informative. \\[3pt]

Avoid crude primitive-only solutions for visually rich concepts, missing key subject details, flat single-block appearances, generic reusable icons that do not reflect the query, readable text labels as semantic shortcuts, and external images, scripts, animation, \texttt{foreignObject}, or remote resources. \\[5pt]

\textbf{Stage 1: Ideal SVG Generation.}
First, imagine a strong visual solution for the query and generate one ideal SVG. The SVG must communicate the query meaning without using readable text labels. It should make the main subject and important components visually clear, contain enough subject shape complexity for the depicted concept, include key internal details, show visible layering or part separation when appropriate, and remain clean and renderable without unnecessary clutter. \\[5pt]

\textbf{Stage 2: Rubric Generation.}
Generate a rubric for evaluating other rendered SVG candidates for the same query. The judge model will later see only the user query, the rubric, and the candidate rendered image, but not the ideal SVG. Therefore, the rubric must be fully self-contained and designed for continuous reward scoring in reinforcement learning. Weak outputs should receive low scores, mediocre outputs should receive middling scores, good outputs should score well but not automatically near-perfect, and only truly strong outputs should score near 1.0. A merely recognizable image should not score highly on most items. \\[5pt]

\textbf{Fixed Rubric Items and Weights.}
Generate exactly five rubric items in this fixed order and with these fixed weights: \\[2pt]
1. Overall Prompt Readability -- weight 5 \\
2. Major Component Legibility -- weight 5 \\
3. Subject Shape Complexity -- weight 5 \\
4. Key Internal Detail Coverage -- weight 5 \\
5. Visual Layering -- weight 4 \\[3pt]

Do not add extra items. Do not introduce catch-all items such as ``overall quality'', ``overall beauty'', ``style appeal'', or ``visual polish''. Each item must focus on exactly one distinct dimension. \\[5pt]

\textbf{Item Definitions.} \\[2pt]

\textbf{1. Overall Prompt Readability:}
Focus only on whether a viewer can understand the overall intended meaning of the query by looking at the image. Do not turn it into a checklist of parts, and do not discuss visual complexity, internal details, or layering. \\[2pt]

\textbf{2. Major Component Legibility:}
Focus only on whether the major objects, important parts, actions, states, or object relationships required by the query are clearly readable. Do not discuss subject shape complexity, internal detail richness, or layering. \\[2pt]

\textbf{3. Subject Shape Complexity:}
Focus only on whether the main subject is represented with enough shape complexity for the concept. Penalize candidates that reduce a visually richer subject into an overly simple primitive or low-information geometric substitute. Do not discuss internal detail coverage or layering. \\[2pt]

\textbf{4. Key Internal Detail Coverage:}
Focus only on whether the most important internal visual details of the main subject are present and readable. Reward identity-defining details that make the subject feel complete and informative. Do not discuss outer shape complexity or layering. \\[2pt]

\textbf{5. Visual Layering:}
Focus only on whether the image shows visible layering, separation, or depth of parts instead of appearing like a flat single-layer block. This may include overlap, foreground/background separation, internal region separation, nested shapes, tonal regions, or visibly separated subject parts. Do not discuss semantic correctness, subject shape complexity, or internal detail coverage. \\[5pt]

\textbf{Orthogonality Requirements.}
The five items must be as orthogonal as possible: each item should target one distinct observable dimension, be independently judgeable, and avoid using evidence from one item to justify another. Item 1 and Item 2 must not collapse into the same criterion; Item 3 and Item 4 must not collapse into the same criterion; and Item 4 and Item 5 must not collapse into the same criterion. Do not use item titles such as ``X and Y'', ``complex and detailed'', ``clear and layered'', or ``readable and polished''. \\[5pt]

\textbf{Strict Scoring Guide Requirement.}
For each rubric item, provide a strict item-specific \texttt{scoring\_guide}. The scoring guide must prevent inflated base scores and create meaningful differences among rollout candidates. Each \texttt{scoring\_guide} must be a JSON object with exactly these keys: \texttt{"1.0"}, \texttt{"0.7-0.9"}, \texttt{"0.4-0.6"}, \texttt{"0.1-0.3"}, and \texttt{"0.0"}. \\[3pt]

Do not describe score ranges using vague satisfaction words such as ``partially satisfies'', ``mostly satisfies'', ``adequate'', ``acceptable'', ``decent'', ``good enough'', ``well represented'', or ``looks good''. Instead, describe concrete visible thresholds for each score range. \\[5pt]

\textbf{Scoring Calibration.}
A score of 1.0 is reserved for exceptional, complete, unambiguous, and non-generic fulfillment of the item. Scores of 0.7--0.9 require strong item-specific visual evidence and, for visual items, multiple non-redundant structures or details. Scores of 0.4--0.6 are for mediocre outputs with real but limited evidence, such as recognizable but overly simple, incomplete, weakly detailed, or visually shallow drawings. Scores of 0.1--0.3 are for weak hints, vague resemblance, isolated marks, primitive symbols, or ambiguous evidence. A score of 0.0 is used for absence, contradiction, or total unreadability. \\[3pt]

A crude but recognizable drawing should not exceed 0.6 on most visual items. A primitive-only drawing of a complex concept should not exceed 0.6 on Subject Shape Complexity. A drawing missing identity-defining internal details should not exceed 0.6 on Key Internal Detail Coverage. A flat single-block drawing with no visible separation should not exceed 0.6 on Visual Layering. A vague or generic image that only weakly resembles the query should usually be scored in 0.1--0.3 for semantic items. \\[5pt]

\textbf{Writing Rules.}
For each rubric item, write a short, natural, sample-specific title, exactly one sentence for the description, and a strict scoring guide object. Keep the rubric open-ended rather than reconstructive. Do not require exact color, orientation, placement, count, or geometry unless the query explicitly requires it. Do not write generic scoring guides that could apply unchanged to any SVG. \\[5pt]

\textbf{Output Format.}
Return only two Markdown code blocks in this exact order: first, the ideal SVG in an XML code block; second, the rubric in a JSON code block. Do not output prose outside the two code blocks. The XML block must contain exactly one standalone SVG, starting with \texttt{<svg} and ending with \texttt{</svg>}. The JSON block must be a JSON array of exactly five rubric items. Each rubric item must contain exactly the keys \texttt{title}, \texttt{weight}, \texttt{description}, and \texttt{scoring\_guide}. Do not include the fixed text-penalty item; it will be added later by the script if needed. \\[5pt]

\textbf{Input query:} \texttt{\{description\}}
\end{prompt}

\begin{prompt}{Fixed Text-Hint Penalty Item}
\textbf{Title:} Prompt Text Hint \\[3pt]

\textbf{Weight:} $-2$ \\[3pt]

\textbf{Description:}
The candidate uses readable text, letters, numbers, or labels as prompt-related hints, revealing the intended meaning in text instead of expressing it through the image itself. \\[5pt]

\textbf{Scoring Guide.} \\[2pt]
\textbf{0.9--1.0:} Clearly readable text directly reveals the prompt meaning, acting as a semantic shortcut. \\[2pt]
\textbf{0.7--0.8:} Partially readable text or symbols are visible and plausibly related to the prompt, but are small, ambiguous, or decorative. \\[2pt]
\textbf{0.4--0.6:} Faint text-like marks or letter-shaped geometric patterns appear, but no clearly readable characters can be identified. \\[2pt]
\textbf{0.1--0.3:} A few isolated straight strokes coincidentally resemble letter strokes, with no coherent textual content. \\[2pt]
\textbf{0.0:} No readable text, letters, numbers, or label-like marks are visible anywhere in the image.
\end{prompt}